\pdfoutput=1

\documentclass[11pt]{article}

\usepackage[final]{acl}

\usepackage{times}
\usepackage{latexsym}

\usepackage[T1]{fontenc}

\usepackage[utf8]{inputenc}

\usepackage{microtype}

\usepackage{inconsolata}

\usepackage{graphicx}

\usepackage{enumitem}
\usepackage{subfig}
\usepackage{amsmath,amssymb}
\usepackage{algorithm}
\usepackage[noend]{algpseudocode}

\usepackage{pifont}%
\newcommand{\cmark}{\ding{51}}%
\newcommand{\xmark}{\ding{55}}%
\newcommand{\kinda}{\ding{108}}%
\usepackage{verbatim}

\newcommand{\given}[1][]{\:#1\vert\:}
\DeclareMathOperator*{\argmin}{arg\,min}
\newcommand*{\affmark}[1][*]{\textsuperscript{#1}}

\title{Multi-Level Explanations for Generative Language Models}

\author{Lucas Monteiro Paes\thanks{Denotes equal contribution in alphabetical order.}\affmark[1], Dennis Wei\footnotemark[1]\affmark[2], Hyo Jin Do\affmark[2], Hendrik Strobelt\affmark[2], Ronny Luss\affmark[2],\\
  {\bf Amit Dhurandhar\affmark[2], Manish Nagireddy\affmark[2], Karthikeyan Natesan Ramamurthy\affmark[2],}\\
  {\bf Prasanna Sattigeri\affmark[2], Werner Geyer\affmark[2], Soumya Ghosh\thanks{Work done while at IBM Research.}\affmark[3]} \\
  \affmark[1]Harvard University \hspace{1cm} \affmark[2]IBM Research \hspace{1cm} \affmark[3]Merck Research Labs\\
  \texttt{lucaspaes@g.harvard.edu}, \texttt{dwei@us.ibm.com} %
  }

\begin{document}
\maketitle
\begin{abstract}
Despite the increasing use of large language models (LLMs) for context-grounded tasks like summarization and question-answering, understanding what makes an LLM produce a certain response is challenging. We propose Multi-Level Explanations for Generative Language Models (\texttt{MExGen}), a technique to provide explanations for context-grounded text generation. \texttt{MExGen} assigns scores to parts of the context to quantify their influence on the model's output. It extends attribution methods like LIME and SHAP to LLMs used in context-grounded tasks where (1) inference cost is high, (2) input text is long, and (3) the output is text.  We conduct a systematic evaluation, both automated and human, of perturbation-based attribution methods for summarization and question answering. The results show that our framework can provide more faithful explanations of generated output than available alternatives, including LLM self-explanations. %
We open-source code for \texttt{MExGen} as part of the ICX360 toolkit: \url{https://github.com/IBM/ICX360}.
\end{abstract}

\section{Introduction}
Large language models (LLMs) are being deployed to generate text used for decision-making, e.g., summarizing meetings \cite{Laskar2023BuildingRM}, extracting key points from legal documents \cite{Kanapala2017TextSF}, and answering doctors' questions \cite{xiong-etal-2024-benchmarking}. In these applications, the LLM is grounded in \emph{context} (e.g., legal documents) provided as part of the input. Given an LLM response, users may wish to know which parts of the context were responsible for the response, or whether the response is grounded at all \cite{Huang2023ASO}. %
We propose \textbf{M}ulti-level \textbf{Ex}planations for \textbf{Gen}erative Language Models (\texttt{MExGen}) to fulfill this user necessity by providing explanations for context-grounded text generation.

\texttt{MExGen} computes explanations by comparing multiple model predictions given perturbed versions of the input context. The explanations take the form of attribution scores assigned to parts of the input, quantifying the effect of each part on the model output. %
Accordingly, we also refer to this type of explanation as \emph{input attribution}. \texttt{MExGen} generalizes popular perturbation-based explanation methods such as LIME \cite{ribeiro2016why} and SHAP \cite{NIPS2017_7062} to generative LLMs. Such methods are widely used for text classification \cite{chen2020generating,kim2020interpretation,mosca-etal-2022-shap,ju2023hierarchical}, %
but their application to generative LLMs is still limited (Section~\ref{sec:relWork}).

Perturbation-based input attribution for generative LLMs presents challenges related to having (1) text outputs, (2) high inference cost, and (3) long text inputs. \texttt{MExGen} provides a framework to address these challenges. The framework can be instantiated with different attribution algorithms, and we do so using 
LIME- and SHAP-like algorithms.

\begin{figure}[t]
    \centering
    \includegraphics[width=\linewidth]{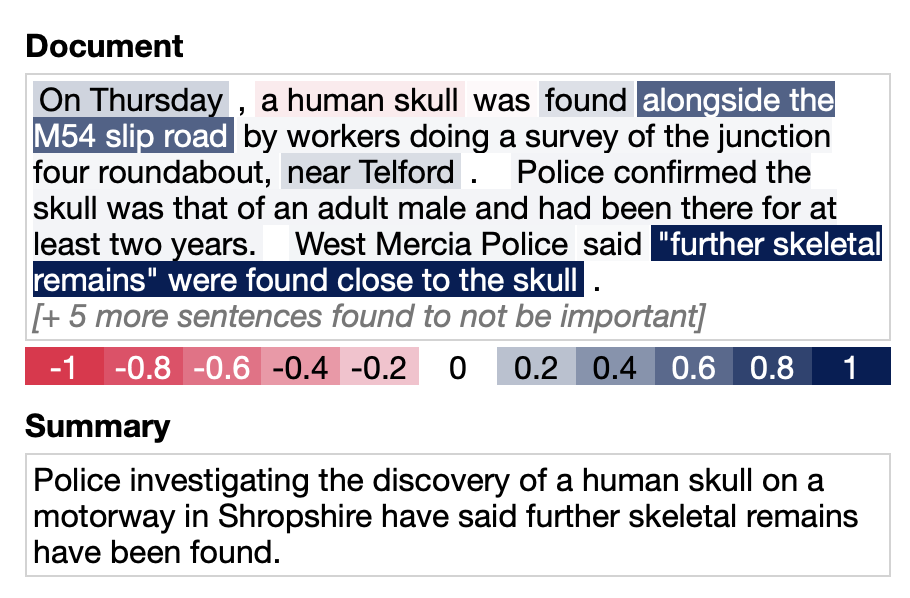}
    \label{fig:teaser}
    \captionsetup{skip=-.5em}
    \caption{Explanation generated using \texttt{MExGen C-LIME} (with \textit{BERT} scalarizer) for a summarization example. The most important phrases found by \texttt{MExGen} (darker blue) suggest that the LLM closely paraphrases text (\textit{further skeletal remains}) and also abstracts concepts (\textit{M54} $\rightarrow$ \textit{motorway}).}
    \vspace{-0.3cm}
\end{figure}

\paragraph{Challenge of output text} The first challenge is that LLMs output text rather than a real number (e.g., the predicted log probability of a class). Attribution algorithms require a real-valued function to quantify that function's sensitivity to different inputs. We address this through the concept of \emph{scalarizers}, functions that map output text to real numbers. We investigate and compare
multiple scalarizers to provide guidance on this choice.

Importantly, most scalarizers that we consider address the truly ``black-box'' setting in which we receive only text as output from the model. This is common with LLMs that only provide API access or are %
proprietary. %

\paragraph{Challenge of input length and LLM inference} %
The input text can be long for context-grounded generative tasks, e.g., whole papers or news articles. %
Generating explanations for long inputs requires more model inferences, imposing higher computational and financial costs. Long inputs also pose interpretation issues, since attributions that are too fine-grained may be less interpretable to a user. %

We address this challenge in %
three ways:
\noindent\textit{(1) Linguistic segmentation:} We segment the input text into linguistic units at multiple levels, for example sentences, phrases, and words. %
\noindent\textit{(2) Multi-level attribution:} We use a refinement strategy that proceeds from attributing at a coarser level like paragraphs to a finer level like words, only refining the most important parts of the context. This controls the number of model inferences and resembles a binary search.
\noindent\textit{(3) Linear-complexity algorithms:} We instantiate our framework with attribution algorithms whose numbers of model inferences scale linearly with the number of units (e.g., number of sentences). %
We also propose a linear-complexity variant of LIME called C-LIME that %
limits the number of units perturbed at one time. %

\paragraph{Evaluation} We evaluate instantiations of \texttt{MExGen} on summarization and context-grounded QA tasks.
We show that \texttt{MExGen} provides more %
faithful explanations compared to baselines (including LLM self-explanation), assigning higher importance to input parts that have the greatest effect on the model output.
We also find that certain scalarizers yield similar explanations to each other. %
Some scalarizers that only use text %
generate explanations close to %
those that depend on log probabilities. %
Human evaluation corroborates the automated evaluation findings and also indicates that %
certain scalarizers and attribution methods, previously considered similar to others in the automated evaluation, were perceived as more faithful by users.

\noindent Our \textbf{main contributions} are: %
\setlist{nolistsep}
\begin{itemize}[noitemsep,leftmargin=4mm]
    \item We propose the \texttt{MExGen} framework to extend perturbation-based input attribution to generative language models, with a multi-level strategy to combat the challenges of long inputs.
    \item We compare several scalarizers for mapping output text to real numbers, notably handling the case of text-only output from the model.
    \item We conduct a systematic evaluation, both automated and human, of input attribution methods for summarization and QA, showing that \texttt{MExGen} can provide more faithful explanations of generated output than the available alternatives. This advantage extends to self-explanations of LLMs 
    \citep{huang2023largelanguagemodelsexplain,madsen-etal-2024-self}, even from powerful LLMs such as DeepSeek-V3. 
\end{itemize}

\begin{figure*}[t]
    \centering
    \subfloat{
        \includegraphics[width=.985\textwidth]{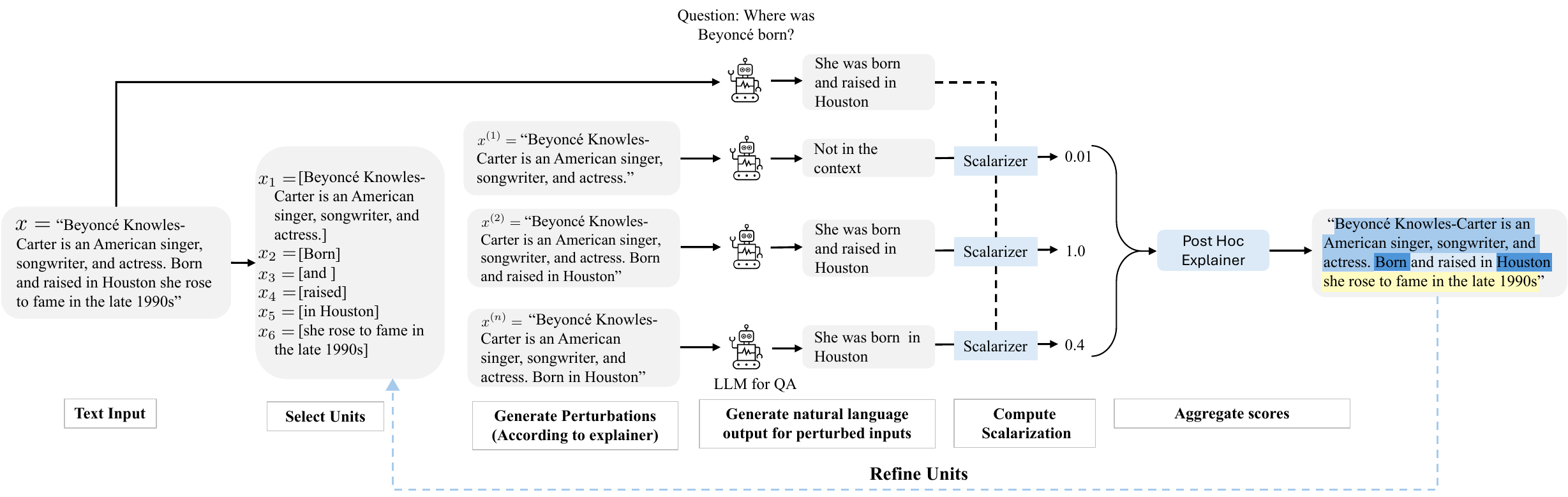}
    }
    \caption{Diagram showing the workflow of \texttt{MExGen}.}
    \label{fig:workflow}
\end{figure*}

\section{Related Work}
\label{sec:relWork}

We discuss works on perturbation-based explanations and self-explanations for \emph{generative} LMs\footnote{Please see Appendix~\ref{sec:appendix_relWork} for other forms of explanation.}. The literature on the former for generative LMs
is limited, as corroborated by \citet{mosca-etal-2022-shap}.

\paragraph{Perturbation-based methods}

\texttt{PartitionSHAP} in the SHAP library handles long inputs by dividing them into token spans and assigning the same score to each token in a span. 
\texttt{PartitionSHAP} produces separate attributions for each output token. This approach is less interpretable because it requires the selection of an output token and assigns multiple attribution scores to each input span. Although \texttt{PartitionSHAP} supports API-only access \cite{shap_summ_example}, understanding its operation required substantial effort (see Appendix~\ref{sec:appendix_scalarizers}).

{\texttt{CaptumLIME}}  
is %
 a modification of LIME \cite{ribeiro2016why} tailored for text generation tasks
in the Captum library \cite{miglani2023using}.\footnote{Captum also has variants of SHAP but we found them slow to run and obviated by %
\texttt{PartitionSHAP}.} 
\texttt{CaptumLIME} allows the user to manually define units for attribution within the input. %
It handles text outputs by computing the log probability of the output. %
This does, however, require access to output probabilities, so \texttt{CaptumLIME} is not suitable for the text-only setting.

\texttt{TextGenSHAP} \cite{textgenshap} offers a more efficient sampling strategy for Shapley value estimation based on speculative decoding \cite{leviathon23}. As with \texttt{CaptumLIME}, such computations also require access to output probabilities. Also, \texttt{TextGenSHAP} is tailored to SHAP. %

Table~\ref{tab:MethodsComp} shows that the above methods each lack capabilities that are offered by \texttt{MExGen}.

\paragraph{Self-explanation methods} These methods prompt the model to explain its predictions (e.g., by ranking the most important parts of its context) \cite{Camburu, huang2023largelanguagemodelsexplain, kroeger2024incontextexplainersharnessingllms, madsen-etal-2024-self}.
However, prior work has found that these self-explanations may be less faithful when used for in-context classification tasks \cite{madsen-etal-2024-self, huang2023largelanguagemodelsexplain}. \citet{FragkathoulasLocal} demonstrated that self-explanations are not as faithful as numerical attributions (like \texttt{MExGen}) in identifying keywords for a QA task. We generalize their work to additional generative tasks and to self-explanations in the form of rankings.

\section{Multi-Level Explanations for Generative Language Models}
\label{sec:mexgen}

This section describes the proposed \texttt{MExGen} framework. %
Figure~\ref{fig:workflow} provides an overview of \texttt{MExGen}.

In the setting of perturbation-based input attribution, we are given a generative LM $f$, an input text sequence of interest $x^o$ (left side of Figure~\ref{fig:workflow}, superscript $o$ for ``original''), and a generated output $y^o = f(x^o)$ that is also a text sequence (top center of Figure~\ref{fig:workflow}). Our goal is to explain $y^o$ (the \emph{target output} for explanation) by attributing to parts of the input $x^o$. Each part of the input, denoted $x_s$, $s = 1,\dots,d$, is to be assigned an attribution score $\xi_s$ (represented by color on the right of Figure~\ref{fig:workflow}) quantifying the importance of $x_s$ in generating the output, in the sense that if important parts are perturbed, then the output will change significantly. As the second through fourth paths in Figure~\ref{fig:workflow} indicate, %
model $f$ can be queried on perturbations $x$ of $x^o$, %
with no further access to $f$.

Generative language tasks pose two main challenges: having text as output, and potentially long text as input. The following two subsections discuss our solutions to these challenges.

\subsection{Handling Text Outputs}
\label{sec:scalarizers}

Input attribution algorithms such as SHAP and LIME require a real-valued %
function as the object to explain (see Section~\ref{sec:multi-level}, ``Linear-complexity algorithms'' for examples of how this function is used). Since the LM $f$ may only output text, we introduce \textit{scalarizers}, which are functions $S$ (shown as blue boxes in Figure~\ref{fig:workflow}) that map output text back to real numbers. We consider two types of access to $f$: (a) \emph{full logit access}, where $f$ can provide predicted logits for all tokens in its vocabulary, at each position in the output sequence; (b) \emph{text-only} access, where we are limited to text outputs. See  Appendix~\ref{sec:appendix_scalarizers} for why we assume  vocabulary-wide logits for (a).

\paragraph{Full logit access} 
When all logits are available, we use the probability of generating the target output %
$y^o$ as the function to explain. We refer to this as the {\textit{Log Prob}} scalarizer. Given output sequence $y^o$ of length $\ell$ and an arbitrary input sequence $x$, we compute the model's log probability of generating each target output token $y^o_t$ conditioned on $x$ and previous output tokens $y^o_{<t}$. We then average over output tokens to obtain the scalarized output for $x$,
\begin{equation}\label{eqn:prob}
    S(x; y^o, f) = \frac{1}{\ell} \sum_{t=1}^\ell \log p\left(y^o_t \given y^o_{<t}, x; f\right).
\end{equation}
Here the scalarizer $S$ is parameterized by $y^o$ and $f$ since the latter is providing predicted probabilities.

The \textit{Log Prob}~scalarizer generalizes the %
log probability used in explaining text classification. This is seen from \eqref{eqn:prob} by setting $\ell = 1$ (single prediction) and identifying $y^o_1$ with the predicted class.

\paragraph{Text-only access} 
Given only the output text $y = f(x)$ generated from the perturbed input $x$, we consider similarity measures $S(y; y^o)$ between $y$ and the target output $y^o$ as scalarizers. These now depend on $f$ only via composition, i.e., $S(f(x); y^o)$, and are not parameterized by $f$ as in \eqref{eqn:prob}. Appendix~\ref{sec:appendix_scalarizers} has further details on these scalarizers.
\begin{itemize}[leftmargin=3mm]
    \item \textit{Sim}: Cosine similarity between embeddings of $y$ and $y^o$ (e.g., SentenceTransformers embeddings).
    \item \textit{BERT}: BERTScore \cite{Zhang*2020BERTScore:} between $y$ and $y^o$.
    \item \textit{BART}: BARTScore (``faithfulness'' version) between $y$ and $y^o$ \cite{yuan2021bartscore}. %
    Measures the probability of an auxiliary LM $f_{\textit{BART}}$ generating $y^o$ given $y$ as input.
    \item \textit{SUMM}: %
    Similar to the \textit{BART} scalarizer with a summarization model as $f_{\textit{BART}}$ \cite{shap_summ_example}. 
    \item \textit{Log NLI}: Log-odds of entailment given $y^o$ as premise and $y$ as hypothesis, computed using a natural language inference (NLI) model.
\end{itemize}

\subsection{Handling Long Text Inputs}
\label{sec:multi-level}
Some generative language tasks require long input texts, such as in summarization and context-grounded QA.
We address this challenge through a combination of three techniques: segmenting the input into linguistic units at multiple levels, using attribution algorithms with linear complexity in the number of input units, and obtaining attributions in a coarse-to-fine manner.

\paragraph{Linguistic segmentation} 
We segment the input into linguistic units at multiple granularities: %
paragraphs, sentences, phrases, and words (``Select Units'' box in Figure~\ref{fig:workflow}). This approach takes advantage of linguistic and other structure present in the input. For example, the input may already be broken into paragraphs or contain multiple distinct retrieved documents, in which case these paragraphs or documents can form the units at the highest level. %
In contrast, existing methods rely on the model's tokenizer, which can yield units (tokens) that are too fine, or treat the text as a flat sequence of tokens and let the algorithm decide how to segment it \cite{chen2020generating,shap_text_examples}. %

We use tokenization and dependency parsing from \texttt{spaCy} v3.6 \cite{spacy} to segment paragraphs into sentences and words.
To segment sentences into phrases, we implemented an algorithm that uses the dependency parse tree from \texttt{spaCy}. In the first pass, the algorithm recursively segments the tree and its subtrees into phrases that are no longer than a maximum phrase length. In the second pass, some short phrases are re-merged. More details are in Appendix~\ref{sec:appendix_phrase}.

Our framework also allows for units at any level to be marked as not of interest for attribution. Textual elements like punctuation, prompt templates, and system prompts are usually not relevant for attribution and, therefore, marked as not of interest. Ultimately, the user may choose which parts of the input context are not of interest.
Appendix \ref{sec:appendix_params:mexgen} gives more details on the choice of ignored units.

\paragraph{Linear-complexity algorithms} 
Given %
a segmented input with possibly mixed units as in Figure~\ref{fig:workflow}, %
we have the task of attributing to units $x_1,\dots,x_d$ (the ``Post Hoc Explainer'' block in Figure~\ref{fig:workflow}). Here, we consider only perturbation-based attribution algorithms that scale linearly with the number of units $d$ in terms of model queries, to control this cost. We instantiate \texttt{MExGen} with three such methods: leave-one-out (\texttt{LOO}), %
a LIME-like algorithm with further constraints (\texttt{C-LIME}), and Local Shapley (\texttt{L-SHAP}). %
In this work, perturbing a unit means simply dropping it; see Appendix~\ref{sec:future} for further discussion.

\texttt{LOO}: Units $x_1,\dots,x_d$ are perturbed one at a time, yielding corresponding perturbed inputs $x^{(1)},\dots,x^{(d)}$. %
The attribution score for $x_s$ is the %
decrease in scalarizer score due to leaving $x_s$ out: %
$\xi_s = S(x^o; y^o, f) - S(x^{(s)}; y^o, f)$.

\texttt{C-LIME}: We use a linear model that operates on interpretable features $z$ and approximates the model $f$ in the vicinity of original input $x^o$. In our case, the interpretable features $z \in \{0,1\}^d$ correspond to units $x_1,\dots,x_d$, with $z_s = 0$ if unit $s$ is perturbed and $z_s = 1$ otherwise. The linear model is fit using $n$ perturbations $x^{(1)},\dots,x^{(n)}$ of $x^o$, with corresponding interpretable representations $z^{(1)},\dots,z^{(n)}$, and scalarized model outputs $S(x^{(1)}; y^o, f), \dots, S(x^{(n)}; y^o, f)$:%
\begin{multline}\label{eqn:lime}
    \xi = \argmin_w \sum_{i=1}^n \pi\bigl(z^{(i)}\bigr) \bigl( w^T z^{(i)} - S(x^{(i)}; y^o, f) \bigl)^2\\ + \lambda R(w),
\end{multline}
where $\pi(z)$ is a sample weighting function and $R$ is a regularizer. 
The best-fit linear model coefficients yield the attribution scores $\xi$. 

We make two main departures from LIME, in addition to the use of a general scalarizer $S$ as seen in \eqref{eqn:lime}. Firstly, we limit the number of perturbations $n$ to a multiple of the number of units $d$. Since $n$ is the number of samples in the linear regression problem \eqref{eqn:lime} while $d$ is the number of parameters to fit, a ratio $n / d$ of 5 or 10 can yield good results. %
LIME by default sets $n$ to be in the thousands independently of $d$, which can be prohibitive for LLMs. Secondly, we limit the number of units $K$ that can be perturbed simultaneously to a small integer. %
This concentrates the smaller number of perturbations on inputs that are closer to $x^o$, which aligns with work showing that doing so improves the fidelity of attributions \cite{tan2023glime}. LIME by contrast samples the number of units to perturb uniformly from $\{1,\dots,d\}$ \cite{mardaoui2021analysis}. %

\texttt{L-SHAP}:
This variation of SHAP by \citet{chen2019lshapley} limits perturbations to a local neighborhood around a unit of interest, overcoming SHAP's exponential complexity. We further extend \texttt{L-SHAP} by also limiting the number of simultaneously perturbed units (as in \texttt{C-LIME}).
The equation for the SHAP score computation is given in \eqref{eq:L_shap_definition}.
Please see Appendix~\ref{sec:appendix_lshap} for further details.

\paragraph{Multi-level explanations}
The multi-level approach is essential for obtaining fine-grained attributions to the most important units in the context without a drastic increase in the computational cost.
Once we have computed attribution scores at a given level, we refine the input units and repeat the process (feedback path at the bottom of Figure~\ref{fig:workflow}), generating \emph{multi-level} explanations. %
For example, given sentence-level attributions, we obtain word-level attributions for a few sentences and keep attributions at the sentence level for the remainder to avoid introducing too many new units.

The few units to be refined are selected by Algorithm \ref{alg:unit_refinement},\footnote{Users can also manually select the units to be refined.} which refines units with scores larger than a predefined threshold $\phi$ and that are among the $k$ units with the highest scores.
Our multi-level approach decreases explanation cost by avoiding attribution of scores to less relevant finer units.
See Appendix~\ref{sec:appendix_params:mexgen} for the used hyperparameters.

\begin{algorithm}[tb]
\caption{Unit Refinement}
\label{alg:unit_refinement}
\begin{algorithmic}[1]
\Require $x_1, \dots, x_d$ \Comment{Current units}

\ \ $k \leq d$ \Comment{Maximum \# units to refine}

\ \ $\phi \in [-1, 1]$ \Comment{Significance threshold}

\State Compute attribution scores $\xi$ using \eqref{eqn:lime} or \eqref{eq:L_shap_definition}.

\State $\psi \gets 2 \frac{\xi - \min_{s}\xi_s}{\max_{s}\xi_s - \min_{s}\xi_s} - 1$ \Comment{Normalize scores}
\State Compute Top-$k$($\psi$) \Comment{$k$ largest $\psi_s$}
\State \texttt{units\_to\_be\_refined} $\gets \{\}$
\For{ $s \in [d]$}
\If{$\psi_s \geq \phi \textbf{ and } \psi_s \in \text{Top-}k(\psi)$}
\State \texttt{units\_to\_be\_refined.add}($x_s$)
\EndIf
\EndFor
\State \Return \texttt{units\_to\_be\_refined} 
\end{algorithmic}
\end{algorithm}

\section{Automated Evaluation}
\label{sec:eval}
We evaluate \texttt{MExGen} %
on two text generation tasks, summarization and context-grounded QA.
Section~\ref{sec:eval:scalarizers} compares the scalarizers that we have considered and provides guidance on the choice of scalarizer. Sections~\ref{sec:eval:explainers} and~\ref{sec:eval:self-explain} compare \texttt{MExGen} to other attribution algorithms and to LLM self-explanations respectively, showing that \texttt{MExGen} has higher fidelity in terms of identifying input units that are most important to the explained model.

\subsection{Setup}
\label{sec:eval:setup}

\paragraph{Datasets and LMs}
For summarization, we evaluate on the Extreme Summarization (XSUM) \cite{narayan-etal-2018-dont} and CNN/Daily Mail (CNN/DM) \cite{see-etal-2017-get, DBLP:conf/nips/HermannKGEKSB15} datasets. In Sections~\ref{sec:eval:scalarizers} and~\ref{sec:eval:explainers}, we use three LMs: the 306M-parameter DistilBART \footnote{\url{https://huggingface.co/sshleifer/distilbart-xsum-12-6}}, the 20B-parameter Flan-UL2  \cite{Tay2023UL2}, and the Llama-3-8B-Instruct (``Llama-3'') \citep{dubey2024llama3herdmodels} models. We treat the DistilBART model as one with full logit access, enabling use of the \textit{Log Prob} scalarizer. We call Flan-UL2 and Llama-3 through an IBM API \citep{ibm_generative_ai_sdk} that only returns text, thus representing the text-only setting precluding the use of \textit{Log Prob.} We evaluate on the first 1000 test set examples of each dataset for DistilBART and the first 500 for Flan-UL2 and Llama-3.

For QA, we use the Stanford Question Answering Dataset (SQuAD) \cite{rajpurkar-etal-2016-squad} (1000 validation set examples, see Appendix~\ref{sec:appendix_params:datasets}). %
We consider two LMs, Llama-3 through the above API, and the 770M-parameter Flan-T5-Large model\footnote{\url{https://huggingface.co/google/flan-t5-large}} \cite{chung2022scaling}, which we treat as providing output logits.

\paragraph{Attribution algorithms}
We instantiate \texttt{MExGen} with the attribution algorithms discussed in Section~\ref{sec:multi-level}: \texttt{LOO}, \texttt{C-LIME}, and \texttt{L-SHAP}. %
For summarization, we obtain mixed sentence- and phrase-level attributions while for QA, we obtain mixed sentence and word attributions, which are appropriate for SQuAD's shorter, paragraph-long contexts. Choices for scalarizers, segmentation, and algorithm parameters are described in Appendix~\ref{sec:appendix_params:mexgen}. %

We compare against \texttt{PartitionSHAP} %
(P-SHAP) \cite{shap_text_examples} and \texttt{CaptumLIME} %
\cite{miglani2023using}. %
For \texttt{PartitionSHAP}, recall from Section~\ref{sec:relWork} that it %
requires the selection of an output token %
to explain, whereas we are interested in %
the output sequence as a whole. For this reason, we modify \texttt{PartitionSHAP} by summing across all attribution scores, corresponding to different output tokens, that it gives to each input span. %
This is equivalent to explaining the sum of the log probabilities of output tokens because of the linearity of Shapley values, and it corresponds to our \textit{Log Prob} scalarizer. 
For \texttt{CaptumLIME}, since it %
accepts user-defined units for attribution, we provide it with the same %
units used by \texttt{MExGen}. Thus we can directly compare Captum's attribution algorithm (i.e., LIME) with ours, controlling for input segmentation. Captum's default for the target function to explain also corresponds to the \textit{Log Prob} scalarizer. %
Additionally, \texttt{PartitionSHAP} and \texttt{CaptumLIME} take the number of model queries as an input; for a fair comparison, we allow them the greater of the numbers of model calls used by \texttt{MExGen} \texttt{L-SHAP} and \texttt{C-LIME}.  

\paragraph{Metrics}
To measure local fidelity of explanations, we use perturbation curves as in \citet{chen2020generating,ju2023hierarchical}. Given a set of attribution scores, %
perturbation curves rank input units in decreasing importance according to their scores, perturb the top $k$ units with $k$ increasing, and plot the resulting change in some output scalarization measuring how much the perturbed output deviates from the target output. 
For the output scalarization, we select an \emph{evaluation scalarizer} from those in %
Section~\ref{sec:scalarizers}. 
To accommodate \texttt{PartitionSHAP} and \texttt{CaptumLIME} (and slightly favor them), we choose an evaluation scalarizer corresponding to the target function that they use for explanation. %
For DistilBART and Flan-T5-Large, this means the \textit{Log Prob} scalarizer, while for Flan-UL2 and Llama-3, we choose \textit{SUMM} in keeping with \texttt{PartitionSHAP}.
Please see Appendix~\ref{sec:appendix_params:perturbation} for further details. %

\subsection{Scalarizer Evaluation}
\label{sec:eval:scalarizers}

\paragraph{Ranking Similarity Across Scalarizers} 
We first compare scalarizers by measuring the Spearman rank correlation between attribution score vectors \(\xi_S\) and \(\xi_{S'}\)  at the same granularity for all scalarizer pairs $S, S'$. Results with cosine similarity instead of Spearman are in Appendix \ref{sec:appendix_automated_evaluation:scalarizers}, Figure~\ref{fig:cosine_similarity_main}.

\begin{figure}[t]
    \centering
    \subfloat[\texttt{MExGen C-LIME}]{
        \includegraphics[width=0.35\textwidth]{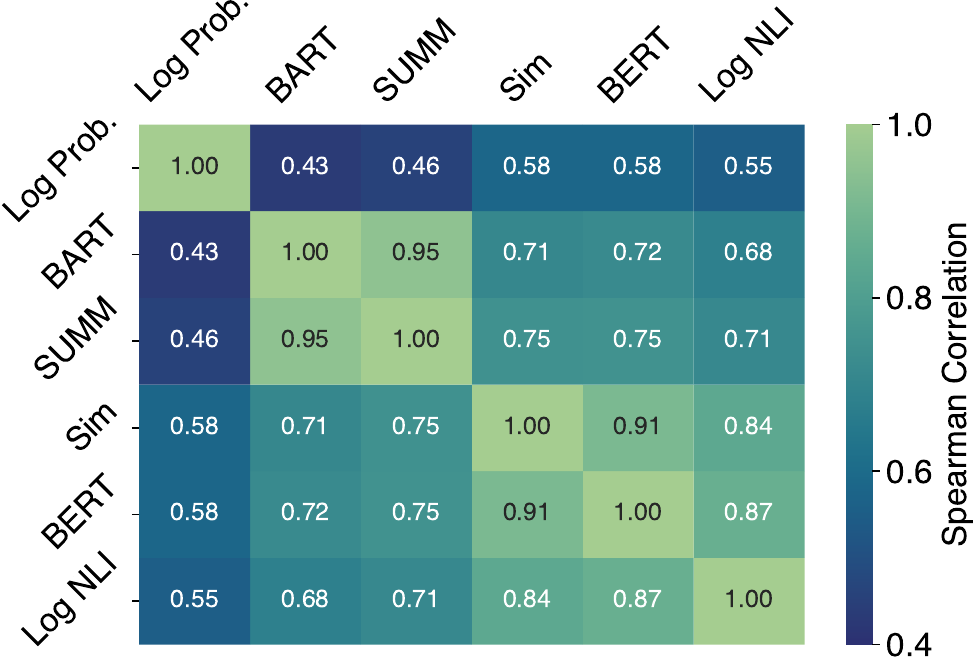}
    \label{fig:spearman_LIME_XSUM}}\\
    \subfloat[\texttt{MExGen L-SHAP}]{
        \includegraphics[width=0.35\textwidth]{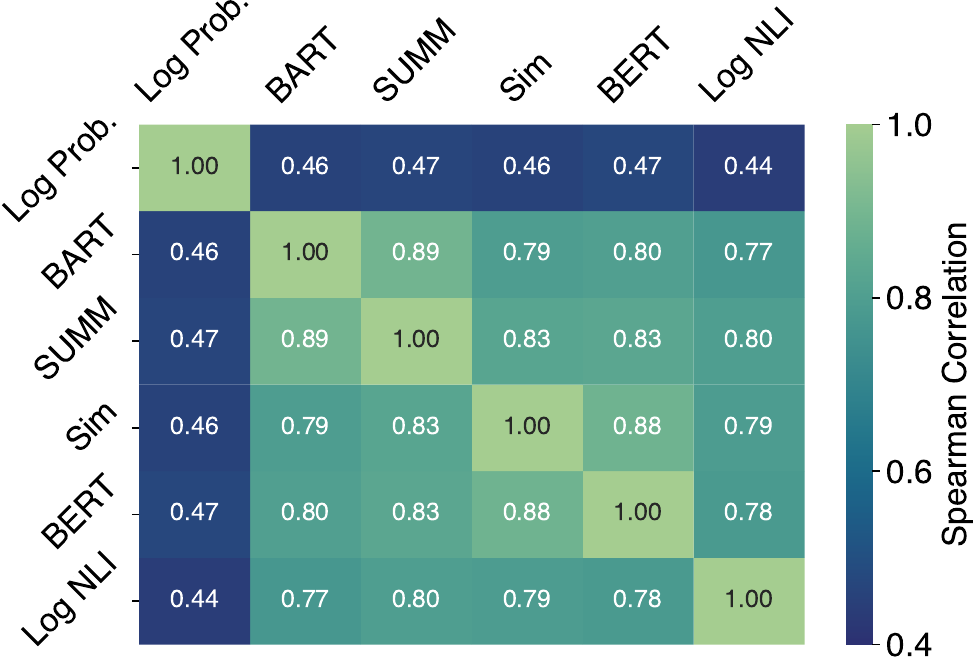}
    \label{fig:spearman_SHAP_SQUAD}}
    \caption{ \label{fig:spearman_main_paper} Spearman's rank correlation between attribution scores using different scalarizers. %
    Attributions were computed using multi-level (a) \texttt{MExGen C-LIME} for the DistilBART model on the XSUM dataset and (b) \texttt{MExGen L-SHAP} %
    for Flan-T5-Large on SQuAD.}
    \vspace{-2mm}
\end{figure}

Figure \ref{fig:spearman_main_paper} shows Spearman correlation matrices, averaged across examples from the respective dataset. Certain scalarizer pairs are highly similar, for example \textit{BART} and \textit{SUMM} as mentioned in Section~\ref{sec:scalarizers}, as well as \textit{Sim} and \textit{BERT}. Thus, it may suffice to consider only one from each pair. \textit{Log Prob}, the only one that uses logits from the model being explained, clearly differs from the others. Additional results in Figure~\ref{fig:cosine_similarity_main} show the same patterns. There are also noticeable differences between Figures~\ref{fig:spearman_LIME_XSUM} and \ref{fig:spearman_SHAP_SQUAD}, which feature different tasks, datasets, and models. For example, Spearman correlation between \textit{BERT} and \textit{SUMM} is $0.75$ in (a) and $0.83$ in (b).  
This suggests %
the necessity of exploring different scalarizers for different tasks.

\begin{figure*}[t]
    \vspace{-1mm}
    \centering
    \subfloat[\textit{Log Prob} as Evaluation]{
        \includegraphics[width=0.31\textwidth]{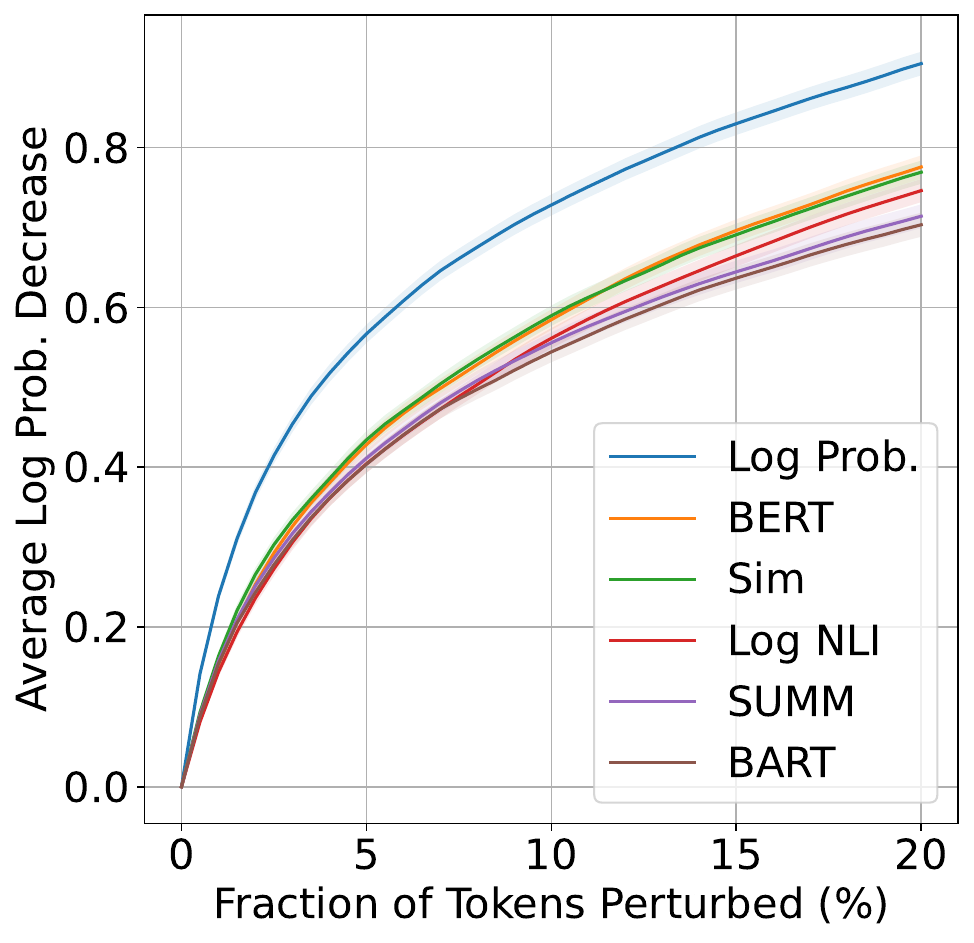}
    \label{fig:SC_LIME_distilbart_xsum_prob}}
    \hfill
    \subfloat[\textit{BERT} Score as Evaluation]{
        \includegraphics[width=0.327\textwidth]{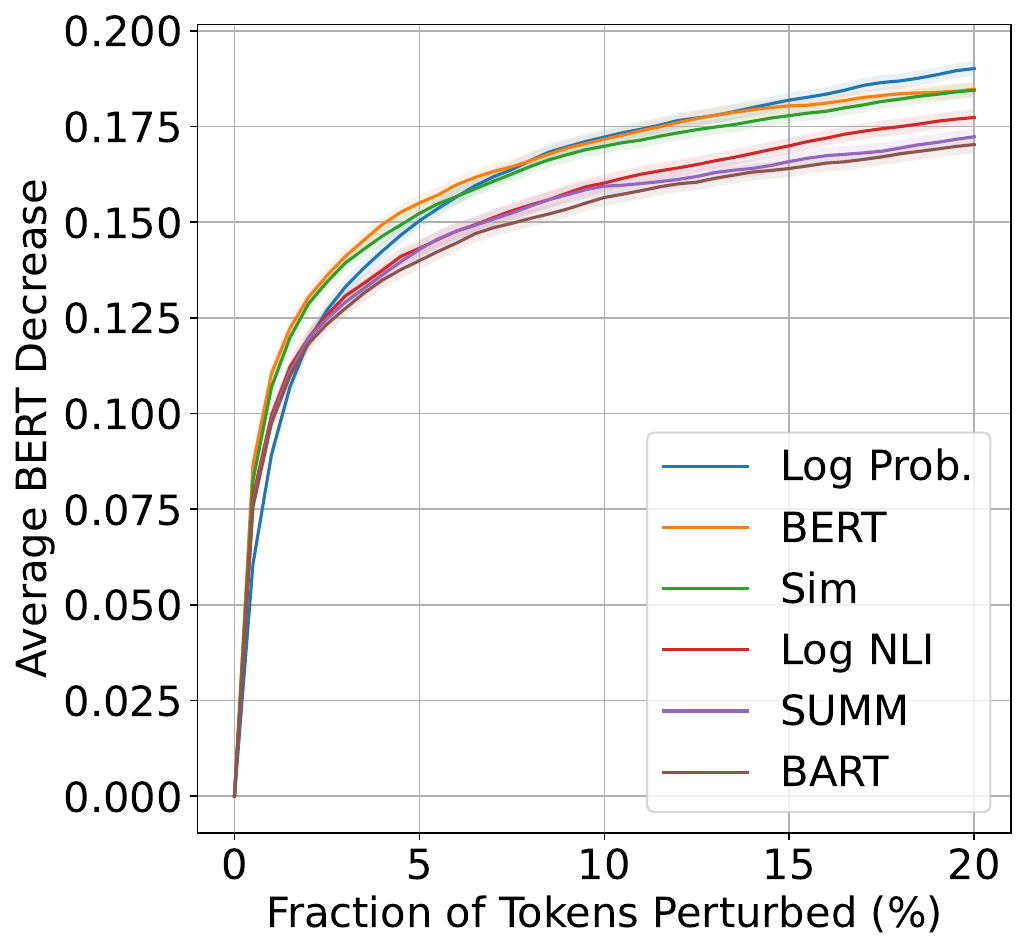}
    \label{fig:SC_LIME_distilbart_xsum_bert}}
    \hfill
    \subfloat[\textit{SUMM} Score as Evaluation]{
        \includegraphics[width=0.31\textwidth]{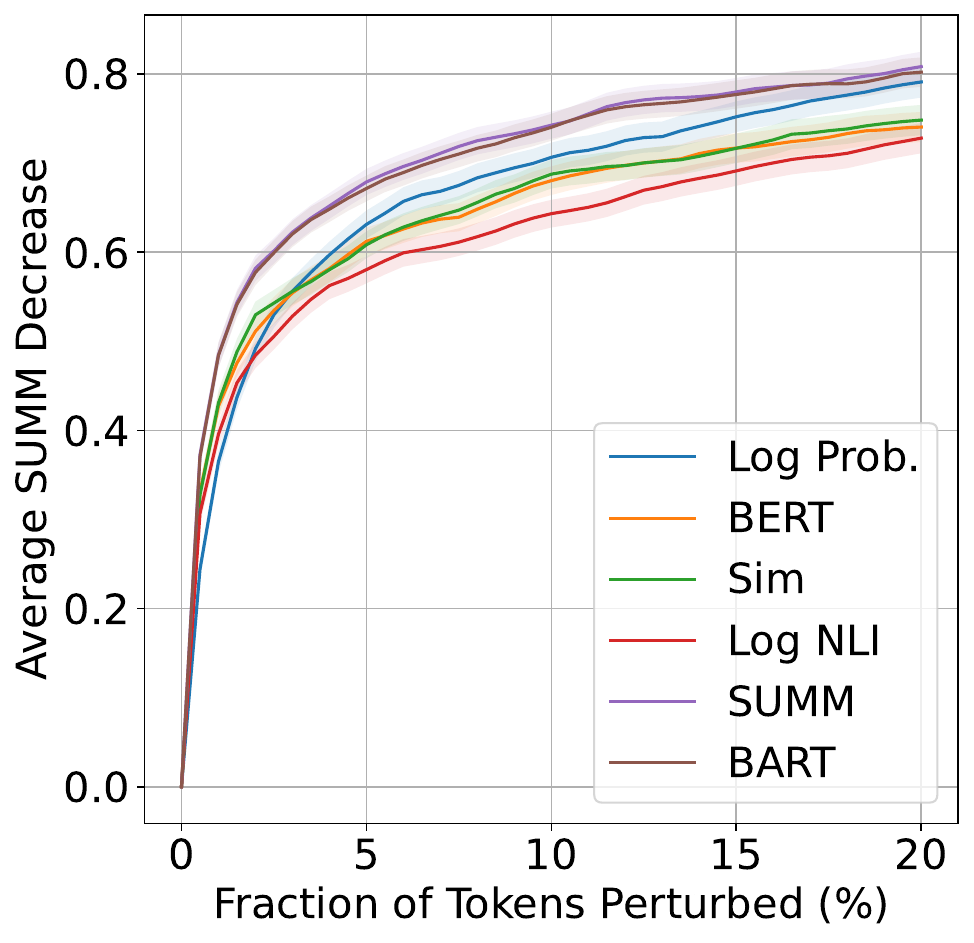}
    \label{fig:SC_LIME_distilbart_xsum_summ}}
    \caption{ \label{fig:ScalarCompLimeBartXSUM} Perturbation curves (higher is better) for \texttt{MExGen C-LIME} with different scalarizers, used to explain the DistilBART model on the XSUM dataset. 
    The curves show the decrease in (a) log probability, (b) BERTScore, (c) \textit{SUMM} score when removing the most important $p$\% of tokens according to each explanation scalarizer. Shading shows standard error in the mean. %
    }
    \vspace{-1mm}
\end{figure*}

\paragraph{Perturbation Curves Across Scalarizers} 
Figures~\ref{fig:SC_LIME_distilbart_xsum_prob}--\ref{fig:SC_LIME_distilbart_xsum_summ} show perturbation curves with different evaluation scalarizers, using \texttt{MExGen} \texttt{C-LIME} with all proposed scalarizers
(i.e., a cross-evaluation of scalarizers).
The curves are averaged over 1000 samples from XSUM (see Appendix~\ref{sec:appendix_params:perturbation} for how) 
and the shading shows one standard error above and below. 
The case where the attribution scalarizer is matched with the evaluation scalarizer is generally the best. %
The \textit{Log Prob} scalarizer performs well across evaluation scalarizers, implying that it gives a more universal explanation (least dependent on the evaluation scalarizer). Hence, we suggest using the log probabilities when they are available.  %

When model logits are not available, %
the choice of scalarizer is not as clear.
However, if (hypothetically) one were to use \textit{Log Prob} as the evaluation scalarizer, %
Figures \ref{fig:spearman_main_paper} and \ref{fig:SC_LIME_distilbart_xsum_prob} indicate that the \textit{BERT} scalarizer %
best approximates the \textit{Log Prob} scalarizer (\textit{Sim} is similar while \textit{SUMM}, the scalarizer used by P-SHAP, performs worse in this regard).
Results for \texttt{MExGen} \texttt{L-SHAP} and a second model-dataset pair in Appendix \ref{sec:appendix_automated_evaluation:scalarizers}, Figures~\ref{fig:ScalarCompXSUMDistilbart}--\ref{fig:ScalarCompShapT5} show similar trends as above.

The computational cost of different scalarizers is discussed in Appendix~\ref{sec:appendix_params:mexgen}.

\subsection{Comparison Between Explainers}
\label{sec:eval:explainers}

\begin{figure*}[t]
    \centering
    \subfloat[DistilBART on XSUM]{
        \includegraphics[width=0.32\textwidth]{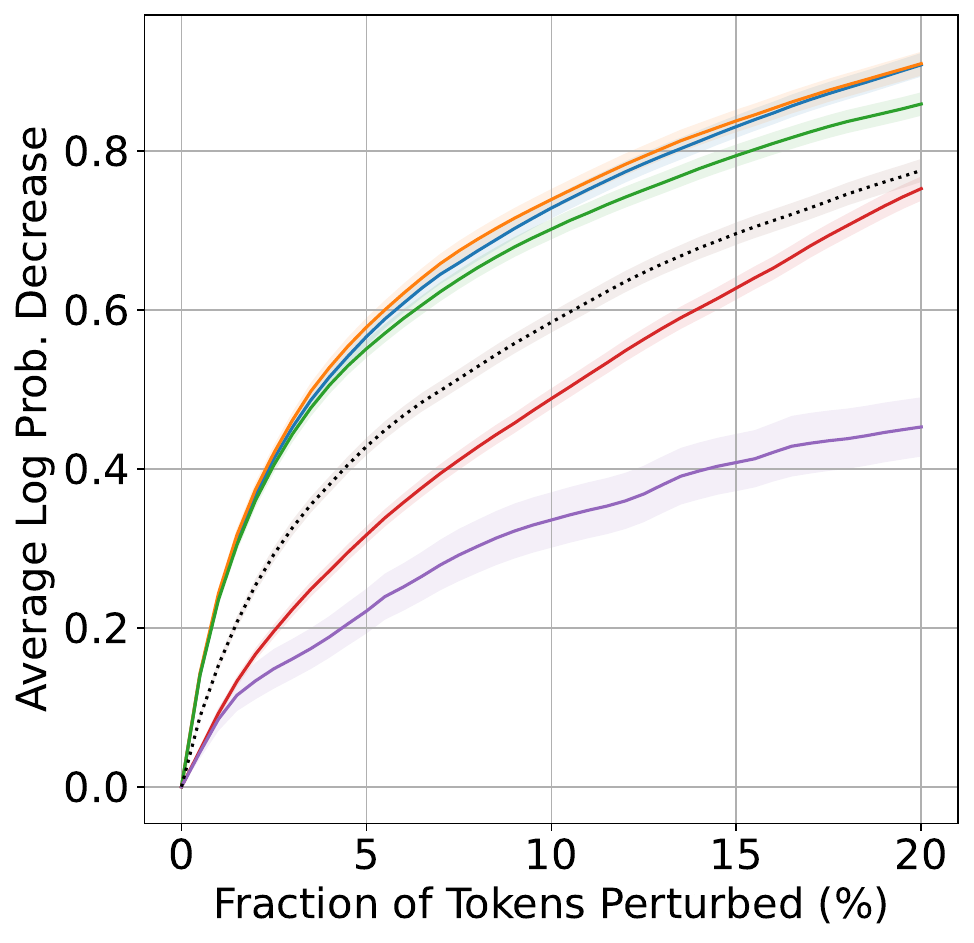}
    \label{fig:Perturbation_distilbart_xsum}}
    \hfill
    \subfloat[Llama-3 on CNN/DM]{
        \includegraphics[width=0.32\textwidth]{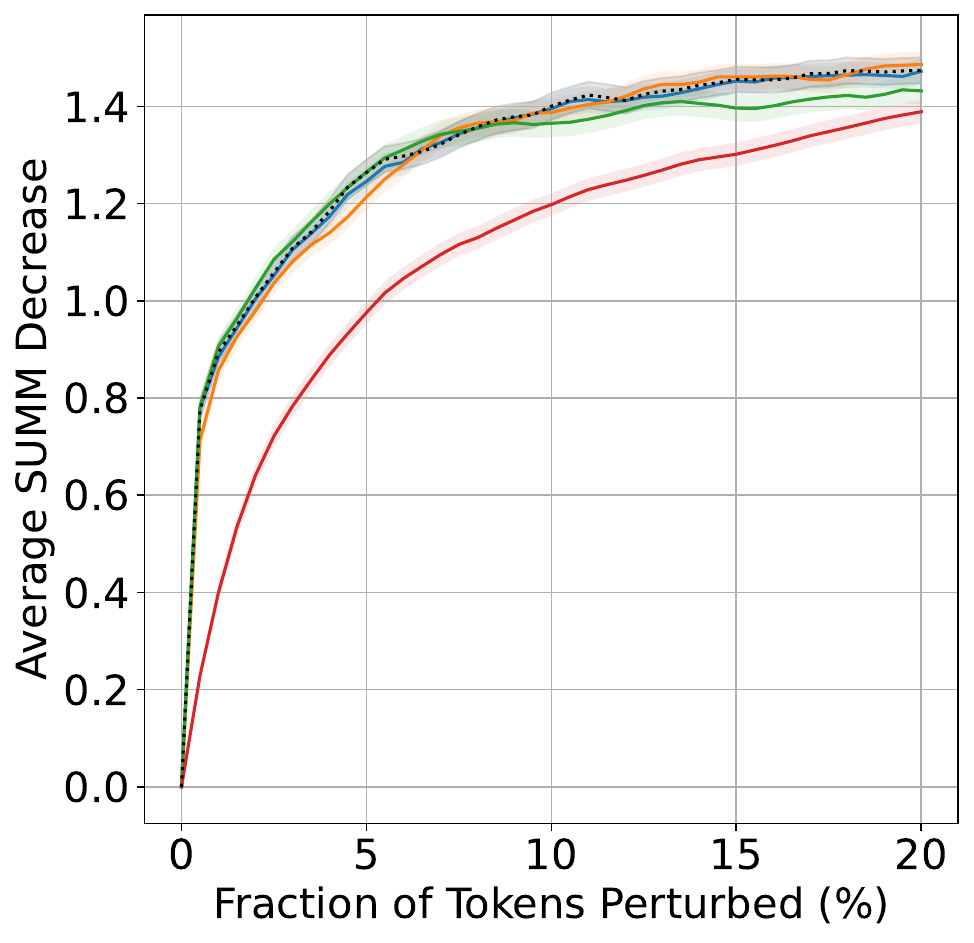}
    \label{fig:Perturbation_llama3-cnndm}}
    \hfill
    \subfloat[Flan-T5-Large on SQuAD]{
        \includegraphics[width=0.32\textwidth]{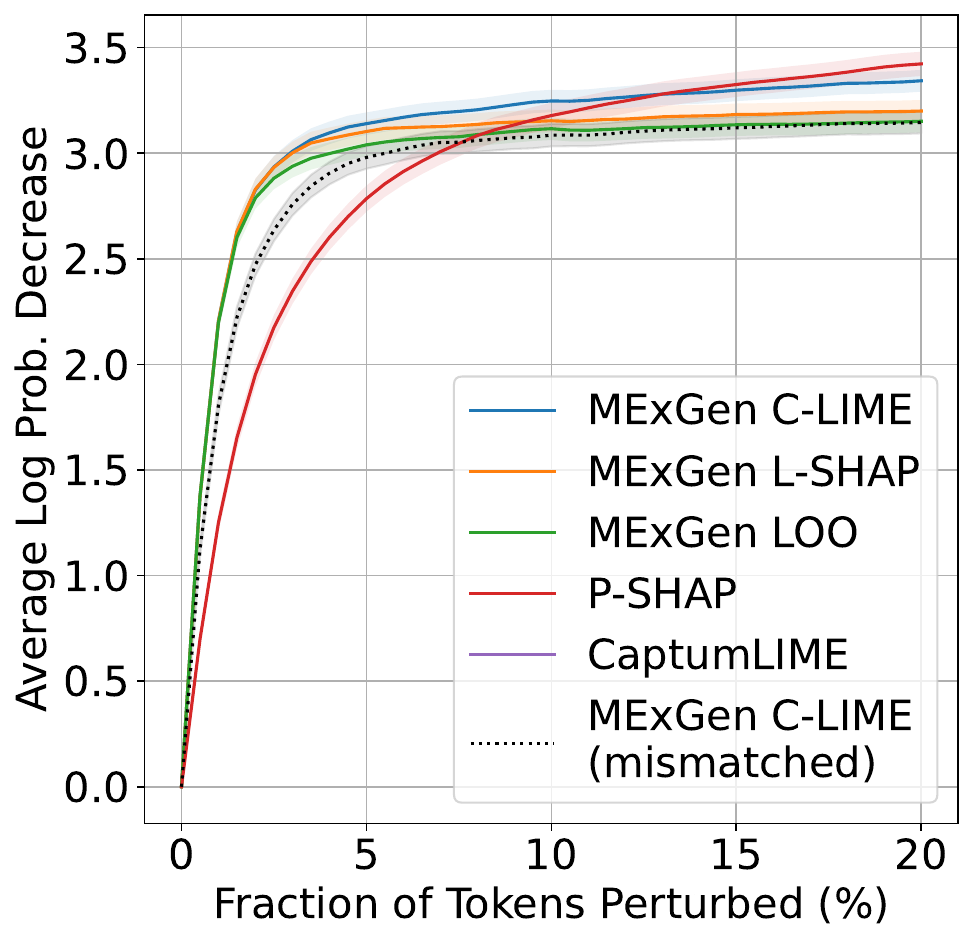}
    \label{fig:Perturbation_flan-t5_squad}}
    \caption{\label{fig:MethodsComparison} Perturbation curves (higher is better) for different explanation methods,
    using the same scalarizer as the evaluation scalarizer in the y-axis label.
    Models and datasets: (a) DistilBART on XSUM, (b) Llama-3 on CNN/DM, (c) Flan-T5-Large %
    on SQuAD. The legend in (c) applies to all panels, but we could run \texttt{CaptumLIME} only for (a). 
    ``\texttt{MExGen} \texttt{C-LIME} (mismatched)'' refers to the use of mismatched scalarizers: BERTScore (a)(c) and BARTScore (b).
    }
\end{figure*}

\paragraph{Perturbation Curves}  
Figure~\ref{fig:MethodsComparison} compares the perturbation curves of \texttt{MExGen} instantiations with \texttt{P-SHAP} and \texttt{CaptumLIME}. Mean curves and standard errors are again computed over the number of examples taken from each dataset (see Section~\ref{sec:eval:setup}). 

Regarding \texttt{CaptumLIME}, we were only able to obtain results for it in Figure~\ref{fig:Perturbation_distilbart_xsum} because (i) we had API access to Llama-3 and \texttt{CaptumLIME} needs output logits, and (ii) \texttt{CaptumLIME} does not support Flan-T5-Large. Figure~\ref{fig:Perturbation_distilbart_xsum} is a direct comparison between LIME (represented by \texttt{CaptumLIME}) and our modification \texttt{C-LIME}, using the same number of model queries and input segmentation as mentioned earlier. \texttt{C-LIME} is clearly more effective.

In comparison with \texttt{P-SHAP}, Figure \ref{fig:MethodsComparison} shows that the perturbation curves for \texttt{MExGen} rise more quickly from zero and are higher for the top $x$\% of tokens, where $x > 20$\% in Figures~\ref{fig:Perturbation_distilbart_xsum} and \ref{fig:Perturbation_llama3-cnndm}, and $x$ varies between 8\% and 13\% in Figure~\ref{fig:Perturbation_flan-t5_squad}. 
This pattern indicates that %
\texttt{MExGen} is better able to identify units that are most important to the model, as measured by the change in the evaluation scalarizer in the leftmost region of each plot. %
Figure~\ref{fig:MethodsComparisonAdd} %
has results on additional model-dataset pairs. Across the tested models and datasets,
\texttt{MExGen} \texttt{C-LIME} and \texttt{L-SHAP} are the top performers, %
while \texttt{MExGen} \texttt{LOO}, the simplest attribution algorithm, is close behind. %

We also show in Figure \ref{fig:MethodsComparison} perturbation curves for \texttt{MExGen} \texttt{C-LIME} using different scalarizers than used for evaluation.
Surprisingly, we find that even when using a mismatched scalarizer (notably \textit{BERT} in Figures~\ref{fig:Perturbation_distilbart_xsum}, \ref{fig:Perturbation_flan-t5_squad}, which does not even use logits), %
\texttt{MExGen} \texttt{C-LIME} can outperform \texttt{P-SHAP} in fidelity. %

\begin{table*}[th]
    \centering
    \small
    \begin{tabular}{ccccccc}
        \hline
        \textbf{Datasets} & \textbf{Models} & \textbf{{MExGen C-LIME}} & \textbf{MExGen L-SHAP} & \textbf{MExGen LOO} & \textbf{P-SHAP} \\
        \hline
        \centering XSUM & DistilBART & \underline{13.6} & \textbf{13.8} & 13.1 & 9.4 \\
        & Flan-UL2 & \underline{17.2} & \textbf{17.4} & 16.7 & 13.7 \\
        & Llama-3-8B-Instruct & \textbf{22.4} & \underline{22.2} & 22.1 & 20.2 \\
        \centering CNN/DM & DistilBART & \underline{13.5} & \textbf{14.7} & 13.2 & 9.7 \\
        & Flan-UL2 & \underline{32.1} & 32.0 & 32.1 & \textbf{33.2} \\
        & Llama-3-8B-Instruct & \textbf{26.4} & \underline{26.3} & 26.1 & 22.1 \\
        SQuAD & Flan-T5-Large & \textbf{62.7} & \underline{61.1} & 60.2 & 58.8 \\
        & Llama-3-8B-Instruct & \underline{56.4} & \textbf{57.0} & 54.9 & 38.5 \\
        \hline
    \end{tabular}
    \caption{ \label{tab:comparing_different_methods}
    Areas under the perturbation curve (AUPC) up to $20$\% of tokens. For DistilBART and Flan-T5-Large, log probability is used as both the explanation and evaluation scalarizer. For Flan-UL2 and Llama-3-8B-Instruct, which do not provide access to log probabilites, \textit{SUMM} is used as the explanation and evaluation scalarizer.
    }
    \vspace{-0mm}
\end{table*}

\paragraph{Area Under the Perturbation Curve} 
Table \ref{tab:comparing_different_methods} shows the area under the perturbation curve (AUPC) to summarize performance over all dataset-model pairs that we tested, including the ones in Figure~\ref{fig:MethodsComparisonAdd}. %
We evaluate AUPC up to $20\%$ of tokens as done in \cite{chen2020generating}.
Across all dataset-model pairs, %
\texttt{MExGen} instantiations (including \texttt{LOO}) performed better (i.e., higher AUPC) than \texttt{P-SHAP}. 
The one exception is the (CNN/DM, Flan-UL2) pair, but even in this case, Figure~\ref{fig:Perturbation_flan-ul2_cnndm} in Appendix~\ref{sec:appendix_automated_evaluation:explainers} shows that the \texttt{MExGen} curves are higher for the top 5\% of tokens.
The second highest AUPC is always from \texttt{MExGen} \texttt{C-LIME} or \texttt{L-SHAP}. %

\paragraph{Computational Cost} 
The computational cost of the compared explanation methods --- measured by time and memory use --- is primarily determined by the number of LLM inferences performed. As noted in Section~\ref{sec:eval:setup}, \texttt{P-SHAP} and \texttt{CaptumLIME} were allocated the same or slightly more inferences than \texttt{MExGen} in our experiments. Even so, \texttt{MExGen} consistently delivered more faithful explanations. %

\subsection{Comparison with LLM Self-Explanation}
\label{sec:eval:self-explain}

\begin{table*}[th]
    \centering
    \small
    \begin{tabular}{cccccccc}
        \hline
        \textbf{Dataset} & \textbf{Model} & \textbf{Scalarizer} & \textbf{{MExGen C-LIME}} & \textbf{MExGen L-SHAP} & \textbf{MExGen LOO} & \textbf{Self} \\
        \hline
        XSUM & Granite-3.3 & Prob & \underline{18.9} & \textbf{19.0} & 18.9 & 9.5 \\
        & & BART & \textbf{15.4} & \underline{14.4} & 14.2 & 12.4 \\
        & DeepSeek-V3 & BART & \textbf{12.7} & \underline{12.3} & 12.3 & 10.5 \\
        CNN/DM & Granite-3.3 & Prob & \underline{17.3} & \textbf{17.4} & 16.9 & 7.1 \\
        & & BART & \textbf{11.9} & 11.0 & \underline{11.0} & 8.9 \\
        & DeepSeek-V3 & BART & \textbf{14.1} & \underline{14.0} & 13.5 & 13.5 \\
        \hline
    \end{tabular}
    \caption{ \label{tab:self-explain}
    Areas under the perturbation curve (AUPC) up to $20$\% of tokens for the comparison with LLM self-explanation. The scalarizer in the ``Scalarizer'' column is used for both explanation and evaluation.
    }
    \vspace{-0mm}
\end{table*}

LLMs are capable of explaining their own outputs, including by providing numerical attributions to their inputs \citep{huang2023largelanguagemodelsexplain,madsen-etal-2024-self}. In this experiment, we compare the fidelity of these self-explanations to that of \texttt{MExGen}. Specifically, since the perturbation curves that we use to evaluate fidelity depend only on the ranking of input units, we prompt the LLM to rank units in order of importance to the output that it generated.

\paragraph{LLMs and Datasets} As a test of current LLMs' ability to self-explain, we chose a large and powerful open-weights LM, DeepSeek-V3. At the time of writing, we could not reliably obtain log probabilities from DeepSeek-V3, so we were unable to apply \texttt{MExGen} with the \textit{Log Prob} scalarizer and could only use a text-only scalarizer. For this reason, we tested a second LLM, Granite-3.3-8B-Instruct, for which both \textit{Log Prob} and text-only scalarizers were possible. We used the summarization datasets XSUM and CNN/DM.

\paragraph{Explanation Methods} For self-explanation, we describe in Appendix~\ref{sec:appendix_params:baselines} the prompt that we used and other details. The most important point is that we made the ranking task easier by following %
\citet{zhang2024longcite} in numbering input units with tags and only asking for a list of tags. %
\texttt{MExGen} was run in the same way as before (see Appendix~\ref{sec:appendix_params:mexgen}).

\paragraph{Results} Table~\ref{tab:self-explain} shows the AUPC values for this self-explanation experiment. The corresponding perturbation curves are in Figure~\ref{fig:self_explain}. In all cases, \texttt{MExGen} is more faithful to the LLM's behavior (as measured by higher AUPC) than the LLM's self-explanations. Considering Granite-3.3, the advantage of \texttt{MExGen} is especially large (AUPC is double or more) when it uses the \textit{Log Prob} scalarizer, and less so with the text-only \textit{BART} scalarizer. The larger DeepSeek-V3 model narrows the gap further on the CNN/DM dataset. Overall, our results indicate that while LLM self-explanations can be good, they are outperformed by algorithms that systematically quantify importance. It is also important to note that this experiment only assesses \emph{ranking} ability, whereas \texttt{MExGen} also provides \emph{real-valued} scores, a task that would be harder for an LLM.

\section{User Study}
We conducted a user study %
to understand how humans perceive explanations provided by different scalarizers and attribution methods, %
and whether they can discern performance differences akin to the quantitative evaluations in Section~\ref{sec:eval}. 
To ease interpretation, we developed a visualization tool that highlights input text spans %
based on a color-coded scale for the attribution scores. 
In this section, we focus on %
the following research questions: %
\begin{enumerate}[leftmargin=5mm]
\item \textit{Fidelity}: Which method is perceived to be better at explaining how the language model generated the summary?
\item \textit{Preference}: Which method do people prefer?
\end{enumerate}
\noindent We discuss additional research questions on concentration and granularity of attribution scores in Appendix~\ref{appendix:user_study}.

We selected ten %
examples from the XSUM dataset for the user study with %
diversity in topics, while ensuring that they do not contain sensitive issues or obvious errors.
We designed an online survey consisting of three parts. Each part showed an input text, randomly drawn from the ten examples, and its summary, generated by the DistilBART model. %
This user study focused on algorithms and scalarizers that are of most interest %
based on the automated evaluation in Section~\ref{sec:eval}. Specifically, we compared two scalarizers (\textit{Log Prob}, \textit{BERT}) %
and three attribution algorithms (\texttt{C-LIME}, \texttt{L-SHAP}, \texttt{PartitionSHAP}). %
The presentation order of the attribution algorithms was randomized to mitigate order effects. The survey consisted of seven pairwise comparisons in total followed by questions for the participants.

We recruited participants from a large technology company who self-identify as machine learning practitioners using language models and collected data from 88 of them after filtering. %
Here, we report a summary of key results only. See Appendix~\ref{appendix:user_study} for details including survey questions, analysis, and statistical results. 

\textbf{Scalarizers.} 
Significantly more participants perceived \textit{BERT} to be higher in fidelity than \textit{Log Prob} (57\% to 35\%). They also preferred \textit{BERT} over \textit{Log Prob} (64\% to 31\%). This result is notable because \textit{BERT} uses only text output from the LM while \textit{Log Prob} depends on output logits. %

\textbf{Attribution methods.}
Significantly more participants perceived \texttt{C-LIME} to be higher in fidelity than \texttt{L-SHAP} ($p$-value $= 0.011$). They also preferred \texttt{C-LIME} over \texttt{L-SHAP} ($p = 0.007$). This result is notable because \texttt{C-LIME} and \texttt{L-SHAP} performed very similarly in the automated evaluation in Section~\ref{sec:eval:explainers}.

\section{Concluding Remarks}
\label{sec:conclusion}
We proposed \texttt{MExGen}, a framework to provide explanations for LLMs used in context-grounded tasks like summarization and question answering.
\texttt{MExGen} uses a multi-level strategy to efficiently explain model predictions in the case of long inputs.
\texttt{MExGen} can produce explanations even when only text outputs are available (API access), thanks to \emph{scalarizers} that map text to numerical values.
Our experiments offer guidance on the choice of scalarizer and show that instances of \texttt{MExGen} provide more faithful explanations, outperforming the baselines %
\texttt{PartitionSHAP} and \texttt{CaptumLIME} as well as self-explanations by powerful LLMs. %
The user study results align with the automated evaluation, and %
reveal that people %
perceive the \textit{BERT} scalarizer as more locally faithful than the \textit{Log Prob} scalarizer. %
This result implies that in some cases, there may be no loss in having text-only access compared to full logit access.

\section*{Limitations}

We see the following limitations and risks:

(1) \texttt{MExGen} is a framework for post hoc explanations.
Although such explanations can help practitioners understand model behavior, they do not fully characterize how models generate output and only provide local explanations. 

(2) %
Although the findings of our automated evaluation are consistent across the tested models and datasets, the results reported in Section \ref{sec:eval} could still change in other experimental settings.

(3) Our user study analyzes the \emph{perception} of participants of how well a method explains the predictions of a model, and not necessarily the fidelity of the explanation itself --- fidelity is measured more directly in the automated evaluation.
However, we believe the fact that the participant pool was composed of people with experience in ML and LLMs improves the quality of their evaluation. 

(4) Post hoc explanations in general come with the risk of being steered to obfuscate undesirable behavior from the model. One potential mitigation is to couple them with additional types of explanation, where possible. Another mitigation is to have a different party compute explanations, rather than the model developer. The black-box, perturbation-based explanations considered in this work lend themselves to such a two-party scenario, where the party computing explanations only needs to query the model to do so. This party would select which perturbed inputs to query the model on, avoiding one path to obfuscation where a model developer uses specially chosen perturbations to conceal undesirable behavior. The explaining party should also be given a large budget of model queries to better probe model behavior.

\section*{Ethics Statement}
\texttt{MExGen} is a framework to explain generative language models. Hence, its objective is to elucidate how a model made a specific prediction.
Methods that aim to understand how black-box models generate their output are essential for guaranteeing transparency during decision-making.
For example, a generative language model can be used to summarize dialog and create minutes of meetings that can later be used to perform high-stakes decisions.
Then, it is necessary to understand how the model generated the summary and ensure that the output content is based on the input dialog.
Therefore, for such high-stakes applications, methods that can provide explanations for text generated by language models are necessary, highlighting the importance of \texttt{MExGen}.

\section*{Acknowledgements}

The authors thank the following people: Inkit Padhi for suggestions on parsing sentences into phrases, using language models for perturbing words, and participation in multiple discussions; Ella Rabinovich and Samuel Ackerman for discussions on and recommendations of similarity measures as scalarizers, and Ella in particular for also providing suggestions on parsing sentences into phrases and using language models for perturbation; Pratap Kishore Varma Vemulamanda for helping to investigate the API-only case of SHAP's summarization example \cite{shap_summ_example}; Subhajit Chaudhury for suggestions of similarity measures as scalarizers; Keshav Ramji for discussion on related work and quality evaluation for the explanations; Kush Varshney and Eitan Farchi for general encouragement and support. 
The authors also thank the anonymous ARR reviewers and action editor for their high-quality reviews and positive, constructive comments, especially Reviewer GEAE for suggesting the comparison with LLM self-explanations, and Yannis Katsis for his advice on the setup for this experiment. 
The work of Lucas Monteiro Paes was supported by the Apple Scholars in AI/ML Fellowship.

\bibliography{custom}

\newpage
\appendix

\begin{table*}[ht]
\centering
\normalsize
\begin{tabular}{cccc}
\hline
\textbf{Method} & \textbf{Long Text Output} & \textbf{Long Text Input} & \textbf{API Access} \\
\hline
\texttt{LIME} & \textcolor{red}{\xmark} & \textcolor{red}{\xmark} & \textcolor{red}{\xmark}\\
\texttt{SHAP} & \textcolor{red}{\xmark} & \textcolor{red}{\xmark} & \textcolor{red}{\xmark} \\
\texttt{HEDGE} & \textcolor{red}{\xmark} & \textcolor{green}{\cmark}  & \textcolor{red}{\xmark} \\
\texttt{P-SHAP} & \textcolor{yellow}{\kinda} & \textcolor{green}{\cmark} & \textcolor{green}{\cmark} \\
\texttt{TextGenSHAP} & \textcolor{red}{\xmark} & \textcolor{green}{\cmark} & \textcolor{red}{\xmark} \\
\texttt{Captum} & \textcolor{green}{\cmark} &  \textcolor{red}{\xmark} & \textcolor{red}{\xmark}\\
\textbf{\texttt{MExGen}} & \textcolor{green}{\cmark} & \textcolor{green}{\cmark} & \textcolor{green}{\cmark}\\
\hline
\end{tabular}
\caption{
Comparing the features of our \texttt{MExGen} framework with existing perturbation-based explanation methods.}
\label{tab:MethodsComp}
\end{table*}

\section{More on Related Work}
\label{sec:appendix_relWork}

Table~\ref{tab:MethodsComp} compares the features of the proposed \texttt{MExGen} framework to alternative perturbation-based explanation methods.
The remainder of this appendix discusses other categories of explanation methods.

\paragraph{Hierarchical Explanations}
A line of work \cite{singh2019hierarchical,Jin2020Towards,chen2020generating,ju2023hierarchical}
has developed hierarchical explanations for sequence models including LLMs, which can reveal compositional interactions between %
words and phrases. %
In particular, the \texttt{HEDGE} algorithm of \citet{chen2020generating} was identified by \citet{mosca-etal-2022-shap} as ``arguably the most suitable choice'' for NLP input attribution, in part because it builds its hierarchy in a top-down, divisive fashion (as opposed to bottom-up agglomeration \cite{singh2019hierarchical,ju2023hierarchical}), which is more practical for long texts. However, \texttt{HEDGE} is specific to classification because it measures feature importance based on classification margin.

\paragraph{Gradient-Based Explanations} Gradient-based methods provide input attribution explanations \cite{simonyan2013deep,sundararajan2017axiomatic,shrikumar2017learning,bach2015pixel}, but they require access to model gradients with respect to the input.

\paragraph{Attribution for Generative Language Models.} Concurrently with our work, other studies have explored attribution techniques for generative LMs \cite{XuhongGilot, cohenwang2024contextcite, sarti-etal-2023-inseq}. 
\texttt{GiLOT} \cite{XuhongGilot} employs optimal transport to compute explanations. However, it focuses on the overall distributional changes across all possible outputs rather than attributing specific generated outputs to particular inputs, as \texttt{MExGen} does. 

\texttt{ContextCite} \cite{cohenwang2024contextcite} extends LIME \cite{ribeiro2016why} to generative models by considering parts of the context as features. While effective, \texttt{ContextCite} operates at a single predefined granularity level (e.g., sentence level), lacking the multi-level nature of \texttt{MExGen}, which improves explanation performance. 

\texttt{Inseq} \cite{sarti-etal-2023-inseq} is a toolkit for attribution of generative LMs, offering methods and visualization tools for context attributions. Its methods focus on computing the importance of each input token for every token in the generated output. However, this approach may not fully capture the influence of larger units in the input, and it also does not explain the entire output. 

A key distinction of \texttt{MExGen} is its ability to function without access to the model's logits by utilizing text-only scalarizers, whereas \texttt{GiLOT}, \texttt{ContextCite}, and \texttt{Inseq} all assume access to these logits.

\paragraph{Self-Explanations \& Numerical Attributions}
Other works have used the generative LM itself to provide explanations in line with subsequent outputs.

An example of self-explanation is the chain-of-thought (CoT) \cite{wei2022chain}, where the model explicitly generates the reasoning in natural language used to produce its output.
However, CoT has issues with stability \cite{turpin2023language} and faithfulness \cite{atanasova-etal-2023-faithfulness, lanham2023measuringfaithfulnesschainofthoughtreasoning} in addition to significant variation in quality \cite{nagireddy2023socialstigmaqa}. Moreover, evaluating the faithfulness of CoT is challenging, and \citet{parcalabescu-frank-2024-measuring} argued that existing faithfulness evaluations actually measure self-consistency. \citet{parcalabescu-frank-2024-measuring} proposed their own measure of self-consistency called CC-SHAP, which adapts SHAP in a way similar to how we adapt P-SHAP. %
CoT is less relevant to our work since it provides explanations in natural language rather than numerical attributions.

Another approach to self-explanation is prompting the model to explain its predictions (e.g., by ranking the most important parts of its context) \cite{Camburu, huang2023largelanguagemodelsexplain, kroeger2024incontextexplainersharnessingllms, madsen-etal-2024-self}.
However, prior work has found that these self-explanations may be less faithful when used for in-context classification tasks \cite{madsen-etal-2024-self, huang2023largelanguagemodelsexplain}.

Recently, \citet{FragkathoulasLocal} employed a modified version of \texttt{MExGen} to identify words in the input context that were essential for the model's prediction in a question-answering task. Then, \citet{FragkathoulasLocal} compared the keywords they identified with the keywords extracted using self-explanation, showing that self-explanations are not as faithful as their method. This is a first step in demonstrating that self-explanations are not as faithful as numerical attributions, as provided by \texttt{MExGen}.
Our work generalizes theirs to additional generative tasks and to self-explanations in the form of rankings. 

\paragraph{Rationalization}
Yet another line of self-explanation methods falls under the class of rationalization methods. \citet{lei_emnlp_2016} propose to simultaneously train a rationale extractor with a predictor. The rationale extractor essentially selects parts of the text to use for prediction that fulfill two criteria: they are interpretable and maintain nearly the same prediction as using the full text. As these parts of the text are interpretable, they offer an explanation for the final prediction. \citet{lei_emnlp_2016} employed regularizations to keep the rationales short while encouraging contiguous text, and a sequence of recent literature has focused on improving these rationales. 

\citet{liu_neurips_2022} propose to share the encoder parameters of the extractor and predictor, with the intuition that the extractor and predictor are both seeking to find the most informative part of the input text. \citet{liu_neurips_2023} takes a different approach and considers removing what is unimportant rather than selecting what is important; they minimize a criterion for dependence of the prediction on the text which essentially detects what parts of the text is not required for prediction. \citet{liu_kdd_2023} use a theoretical analysis of how the Lipschitz constant of the predictor affects the extractor to propose how to control the learning rate for obtaining better rationales. \citet{liu_acl_2023} introduce an ensemble of extractors (multiple extractors with different initializations) into the rationalization training framework and find improved prediction performance. At inference time, only the first extractor of the ensemble is used for interpretability. \citet{liu_icde_2024} further include a predictor on the full text into a regularization term in order to better align predictions on extracted text with those of the full text. Most recently, \citet{liu_neurips_2024} build on \citet{liu_neurips_2023} and propose a loss criterion that attempts to treat spurious correlations in the text as if it were noise.

\section{Further Details on \texttt{MExGen}}
\label{sec:appendix_details}

This appendix provides more details on the \texttt{MExGen} framework that are of a more general nature. For parameter settings and other details specific to our experiments, please refer to Appendix~\ref{sec:appendix_params}.

\subsection{Scalarizers}
\label{sec:appendix_scalarizers}

\paragraph{Vocabulary-wide logits}
The inputs $x$ in our case are perturbed versions of the original input $x^o$. Some of these perturbations however may be significantly different semantically and cause the probability of generating the target output $y^o$ to decrease dramatically. For this reason, the \textit{Log Prob} scalarizer may require access to logits for improbable tokens conditioned on $x$, which can be ensured if logits are available for the entire vocabulary, but not if only the top $k$ logits are provided.

\paragraph{Aggregation for \textit{Log Prob} scalarizer}
As alternatives to the average over the output sequence in \eqref{eqn:prob}, other ways of aggregating include using the sum or taking the product or geometric mean of the probabilities. We choose the average to normalize for sequence length and because log probabilities tend to be a more ``linear'' function of inputs than probabilities.

\paragraph{BERTScore}
BERTScore uses an LM to obtain contextual embeddings for the tokens in $y$ and $y^o$, matches the two sets of embeddings based on cosine similarity, and computes a score to quantify the degree of match.

\paragraph{BARTScore} 
BARTScore is the same as \eqref{eqn:prob} except with $y$ in place of $x$ and an auxiliary LM $f_{\textit{BART}}$ (not necessarily a BART model) in place of the LM $f$ being explained. It thus measures the probability of $f_{\textit{BART}}$ generating $y^o$ given $y$ as input.

\paragraph{\textit{SUMM} scalarizer}
We included this scalarizer to represent our understanding of how the abstractive summarization example \cite{shap_summ_example} in the SHAP library handles the text-only API case. We investigated the code behind this SHAP example, and specifically the \texttt{TeacherForcing} class that it uses. Our understanding of the code is that it obtains proxy log-odds for tokens in the target output $y^o$ by taking log-odds from an auxiliary summarization model $f_{\textit{SUMM}}$, with input $y$ and output set to $y^o$. If we then average the log-odds over the tokens in $y^o$ (similar to \eqref{eqn:prob}), the result is similar to the \textit{BART} scalarizer with $f_{\textit{BART}} = f_{\textit{SUMM}}$ (with a possible discrepancy between log-odds versus log probabilities). Our experiments show that the \textit{BART} and \textit{SUMM} scalarizers are indeed very similar.

\paragraph{\textit{Log NLI} scalarizer}
This scalarizer is based on the intuition that \textit{Log NLI} entailment is a kind of similarity. We use the NLI model to predict the log-odds of entailment given $y^o$ as premise and $y$ as hypothesis, and optionally in the other direction as well. If both directions are used, we take the geometric mean of the two entailment probabilities and then convert back to log-odds. We note that bi-directional entailment has been used to approximate semantic equivalence, for example in \citet{kuhn2023semantic}, but here we use the log-odds scores and not just the predicted labels.

\subsection{Phrase Segmentation}
\label{sec:appendix_phrase}
Here we describe the phrase segmentation algorithm mentioned in Section~\ref{sec:multi-level}. The algorithm starts with the root token of the tree and checks whether each child subtree of the root is shorter (in terms of tokens) than a maximum phrase length parameter. If it is shorter, then the subtree constitutes a phrase, and if it is not, the algorithm is recursively applied to the subtree. The root token of each (sub)tree is also taken to be a phrase. Once the sentence has been recursively segmented into phrases in this manner, a second pass is performed to re-merge some phrases that have become too fragmented, thus controlling the number of phrases which is desirable for computation and interpretation. Specifically, phrases that constitute noun chunks (as identified by \texttt{spaCy}) are merged, and certain single-token phrases are merged with their neighbors. Further notes:
\begin{itemize}
    \item Subtrees of the dependency parse tree are usually contiguous spans of text (in English), but sometimes they correspond to multiple spans. In this case, we treat each span as its own subtree since we wish to have contiguous phrases.
    \item In measuring the token length of a span, we do not count punctuation or spaces.
    \item In merging phrases that fall within a noun chunk, we check conditions that are consistent with noun chunks, for example that there is a root phrase that is a single token (the noun), and that the other phrases are children of the root phrase.
    \item For merging single-token phrases with their neighbors, we use the following criteria:
    \begin{itemize}
        \item The single-token phrase (\emph{singleton}) is a non-leaf phrase (is the parent of other phrases) or a coordinating conjunction.
        \item If the singleton is a coordinating conjunction (e.g.~``and''), the neighbor is a corresponding conjunct (e.g.~``Bob'' in ``Alice and Bob'').
        \item If the singleton is a preposition (e.g.~``to''), the neighbor is a child of the preposition (e.g.~``the store'' in ``to the store'').
        \item If the singleton is of some other type, the neighbor is a leaf phrase and is either adjacent to the singleton or a singleton itself.
        \item The merged phrases do not exceed the maximum phrase length parameter.
    \end{itemize}
\end{itemize}

\subsection{\texttt{L-SHAP}}
\label{sec:appendix_lshap}

This \emph{local} variation of SHAP was proposed by \citet{chen2019lshapley} to decrease the number of model inferences relative to %
SHAP, which requires exponentially many inferences. %
In our context, \texttt{L-SHAP} does so by only perturbing units that are within a constant-size neighborhood of the current unit being attributed to. This makes the number of inferences scale linearly with the number of units. 
More precisely, for unit of interest $s \in [d]$, we consider only the radius-$M$ neighborhood $\mathcal{N}_s^M = \{s - M, \dots, s-1, s+1, \dots, s + M\}$ (truncated to $1,\dots,d$ if necessary) and subsets of $\mathcal{N}_s^M$ with cardinality up to $K$. %
Then the attribution score $\xi_s$ for unit $s$ is given by
{%
\begin{multline}\label{eq:L_shap_definition}
    \xi_s = \frac{1}{K+1} \sum_{A \subseteq \mathcal{N}^{M}_{s} : |A| \leq K} %
    \binom{|\mathcal{N}^{M}_{s}|}{|A|}^{-1} \\ 
    \times \left(S(x^{(A)}; y^o, f) - S(x^{(A \cup \{s\})}; y^o, f)\right),
\end{multline}
}%
where $x^{(A)}$ is a perturbation of $x^o$ in which %
units $j \in A$ are perturbed (just dropped in our work). %

\section{Automated Evaluation Details}
\label{sec:appendix_params}

This appendix documents choices and parameter settings used in the experiments in Section~\ref{sec:eval}. We follow the order of presentation in Section~\ref{sec:eval:setup}: datasets and LM inference, choices for the \texttt{MExGen} framework (Appendix~\ref{sec:appendix_params:mexgen}), baseline attribution methods (Appendix~\ref{sec:appendix_params:baselines}), perturbation curve details, and computing environment.

\subsection{Datasets}
\label{sec:appendix_params:datasets}
The GitHub repository\footnote{\url{https://github.com/EdinburghNLP/XSum}} that enables the XSUM dataset \cite{narayan-etal-2018-dont} to be rebuilt is made available under the MIT license. The CNN/DM dataset \cite{see-etal-2017-get} is made available by HuggingFace\footnote{\url{https://huggingface.co/datasets/abisee/cnn_dailymail}} under the Apache-2.0 license. Both datasets are intended for abstractive summarization, which is how we use them. Both datasets consist of news articles and thus contain the names of individuals. Since these names were reported by the media in the original articles, they are already in the public domain and there is no further issue. 

SQuAD is distributed under the CC BY-SA 4.0 license. The dataset is intended for evaluating reading comprehension by answering questions on the contexts in the dataset, which is how we use it as well. Since the contexts are taken from Wikipedia articles, some of them name individuals, but again these articles are already in the public domain. We selected 1000 validation set examples at random from SQuAD because selecting the first 1000 examples yielded insufficient diversity. %

\subsection{Language model inference}
\label{sec:appendix_params:inference}

\paragraph{DistilBART and Flan-T5-Large} The two models that we treated as providing full logit access, DistilBART and Flan-T5-Large, were downloaded from HuggingFace under the Apache-2.0 license. The DistilBART model was trained on the summarization datasets XSUM and CNN/DM, and we use it for summarization. Flan-T5-Large is a general-purpose LM intended for research on LMs, consistent with our use.

The models were called through their \texttt{.generate} method. For generating the original (target) output (corresponding to the original input), \texttt{max\_new\_tokens} was set to \texttt{None} (i.e., the default). When using the text-only scalarizers, perturbed output texts (corresponding to perturbed inputs) are also generated, and for these, \texttt{max\_new\_tokens} was set to $1.5$ times the number of tokens in the target output. The Log Prob scalarizer computes the log probability of generating the target output, so \texttt{max\_new\_tokens} is not needed in this case. All other hyperparameters were left at default settings (for example greedy decoding was used). 

\paragraph{Flan-UL2 and Llama-3} Flan-UL2 and Llama-3-8B-Instruct are general-purpose LMs. Flan-UL2 is distributed by HuggingFace under the Apache-2.0 license while Llama-3 is distributed under its own Meta Llama 3 Community License.\footnote{\url{https://www.llama.com/llama3/license/}} We however accessed Flan-UL2 and Llama-3 using an LLM API service provided by IBM \citep{ibm_generative_ai_sdk} (no longer in existence). 

For these API calls to Flan-UL2 and Llama-3-8B-Instruct, we used greedy decoding and \texttt{max\_new\_tokens} $=100$. In the case of Llama-3, we used its chat template and provided the following system prompts for summarization and QA respectively: (summarization) ``Summarize the following article in one sentence. Do not preface the summary with anything.'' (QA) ``Please answer the question based on the provided context. Answer with a short phrase or sentence.'' No chat template or system prompt was necessary to prompt Flan-UL2 to summarize.

\paragraph{DeepSeek-V3 and Granite-3.3} The DeepSeek-V3 and Granite-3.3-8B-Instruct models used in the self-explanation experiment of Section~\ref{sec:eval:self-explain} were called using a different LLM API service provided by IBM. Similar to Llama-3, we used greedy decoding, \texttt{max\_tokens} $=100$, the model's chat template, and the same system prompt for summarization above. We also fixed the LLM's random seed so that it would generate the same output given the same input. For generating self-explanations in the form of a list of tags, \texttt{max\_tokens} was increased to 500 to accommodate long lists. For Granite-3.3, we were able to obtain log probabilities by setting \texttt{max\_tokens} $= 0$ and \texttt{logprobs} $= 0$, which returns log probabilities of tokens in the input given to the LLM.

\subsection{\texttt{MExGen}}
\label{sec:appendix_params:mexgen}

\paragraph{Scalarizer models}
The text-only scalarizers presented in Section~\ref{sec:scalarizers} can be instantiated with different models. The ones used in our experiments are as follows:
\begin{itemize}
    \item ``\textit{Sim}'': We use the all-MiniLM-L6-v2 embedding model from the SentenceTransformers package \cite{reimers-2019-sentence-bert}.\footnote{This model can be found at \url{https://huggingface.co/sentence-transformers/all-MiniLM-L6-v2} and is distributed under the Apache-2.0 license.}
    \item ``\textit{BERT}'': We use the model deberta-v2-xxlarge-mnli\footnote{This model can be found at \url{https://huggingface.co/microsoft/deberta-v2-xxlarge-mnli} under the MIT license.} \cite{he2021deberta} to compute BERTScore \cite{Zhang*2020BERTScore:} (MIT license). This is the same model that we use for the \textit{Log NLI} scalarizer. Our initial reason for doing so was to see whether the two scalarizers would be very similar because of this choice (they are not as they operate on different principles). We take the ``F1-score'' output as the BERTScore.
    \item ``\textit{BART}'' and ``\textit{SUMM}'': For ``\textit{SUMM}'', we follow \citet{shap_summ_example} in using the same distilbart-xsum-12-6 model as both the scalarizing summarization model as well as the primary summarization model to explain. For ``\textit{BART}'', i.e., BARTScore \cite{yuan2021bartscore}, we use the code\footnote{\url{https://github.com/neulab/BARTScore}, Apache-2.0 license} from the authors and also instantiate it with the distilbart-xsum-12-6 model. The purpose was to determine whether the ``\textit{BART}'' and ``\textit{SUMM}'' scalarizers are very similar when instantiated with the same model, which is indeed the case.
    \item ``\textit{Log NLI}'': As mentioned above, we use deberta-v2-xxlarge-mnli as the \textit{Log NLI} model. We also choose to compute the \textit{Log NLI} entailment probability in both directions and take the geometric mean of the two before taking the logit.
\end{itemize}
All other parameters of these scalarizers are kept at default values.

In the case where log probabilities are not available from the LLM being explained, the above text-only scalarizers contribute to the overall computational cost of \texttt{MExGen}, but only in a secondary way. This is because our choices of scalarizers are small LMs (at most $1.5$B parameters for deberta-v2-xxlarge-mnli). The \textit{Sim} scalarizer was the most computationally efficient because it uses a SentenceTransformer model. The other scalarizers were more comparable to each other, with the \textit{Log NLI} scalarizer being the most costly because of our choice to compute bi-directional entailment (two inferences versus one).

\paragraph{Linguistic segmentation}
As mentioned in Section~\ref{sec:multi-level}, we use \texttt{spaCy} v3.6 \cite{spacy} (distributed under the MIT license) for sentence and word segmentation. Our custom phrase segmentation algorithm has one main parameter, the maximum phrase length, which we set to $10$ \texttt{spaCy} tokens (not counting spaces and punctuation). 

In terms of units that are not of interest for attribution, we exclude non-alphanumeric units (generally punctuation and newlines) and prompt template elements, for example start-of-message and end-of-message tokens like ``<|start\_header\_id|>user<|end\_header\_id|>'', prefixes such as ``Context: '', and system prompts. For QA, we exclude the question and allow only the context to be attributed to.

\paragraph{Multi-level explanations}
The decision here is how many top sentences are refined into phrases in the summarization experiments and into words in the QA experiments. For this we follow Algorithm~\ref{alg:unit_refinement} with the following parameter settings: For summarization with the DistilBART model, the number of top sentences is set to $k = 3$ and the significance threshold is set to  %
$\phi = 1/3$ for sentence-level scores from \texttt{C-LIME} and \texttt{LOO} and $\phi = 0.3$ for \texttt{L-SHAP}. %
For summarization with Flan-UL2, Llama-3, DeepSeek-V3, and Granite-3.3, we set $k$ to correspond to the top 25\% of sentences, %
rounding %
to the nearest integer and ensuring $k \geq 1$. A significance threshold is not used, i.e., $\phi = -1$. %
For QA, we simply take the top $k = 1$ sentence as the context paragraphs in SQuAD tend to have only a handful of sentences, and again do not use a significance threshold ($\phi = -1$).

\begin{figure*}[th]
    \centering
    \fcolorbox{black}{gray!20}{\begin{minipage}{\linewidth}
    You provided the summary below of an article, also below. The article is divided into units (sentences or phrases), which are numbered in the format: <u0> unit 0 <u1> unit 1 ... Please list the \{\texttt{top\_k}\} units that were most important for you to produce this summary. List them in order from most important to least important. List only the unit numbers, for example ``<u3>, <u1>, <u4>''.\\
    \\
    Summary:\\
    \{\texttt{summary}\}\\
    \\
    Article:\\
    <u0> \{\texttt{units[0]}\} <u1> \{\texttt{units[1]}\} \ldots \\ 
    \end{minipage}}
    \caption{Prompt for LLM self-explanation}
    \label{fig:self_explain_prompt}
\end{figure*}

\paragraph{\texttt{C-LIME}}
For \texttt{C-LIME}, the parameters controlling the perturbations are the constant of proportionality between the number of perturbations $n$ and number of units $d$, and the maximum number of units $K$ perturbed at one time. For the smaller DistilBART and Flan-T5-Large models, $n/d$ is set to $10$ and $K$ to $3$, except for sentence-level attributions on SQuAD where $K = 2$. 

For Flan-UL2, Llama-3-8B-Instruct, DeepSeek-V3, and Granite-3.3-8B-Instruct, since inference is more costly for these larger models, we take the following approach. First, at the sentence level of attribution, we only use \texttt{LOO} to identify the top sentences and do not use \texttt{C-LIME}. At the phrase level for the summarization datasets, we use \texttt{C-LIME} with $K = 2$ and $n$ equal to the total number of phrases (in the entire document, not just in the sentences selected for refinement). Similarly, at the word level for SQuAD, we set $K = 2$ and $n$ equal to the total number of words. In both cases this corresponds to a ratio $n / d \approx 5$. 

\texttt{C-LIME} has two additional aspects to consider:

\textit{Sample weighting:} ($\pi(z)$ in \eqref{eqn:lime}) Since we limit the number of units $K$ that are simultaneously perturbed to a small integer, we also no longer use LIME's sample weighting scheme. Instead, we give each subset cardinality $k = 0,\dots,K$ the same total weight (say $1$ without loss of generality), and then distribute this weight uniformly over the subsets of that cardinality.

\textit{Regularization:} Different regularizers $R(w)$ can be used in \eqref{eqn:lime}, e.g. $\ell_2$ or $\ell_1$. In our experiments however, we do not regularize (i.e., $R(w) \equiv 0$) and compute a fully dense solution. This allows all units to be ranked to facilitate evaluation of perturbation curves. 

\paragraph{\texttt{L-SHAP}}
For \texttt{L-SHAP}, the two main parameters are the local neighborhood radius $M$ and the maximum number of neighbors $K$ perturbed at one time. (Note that $K$ does not include the unit of interest, so altogether the maximum number of perturbed units is $K + 1$.) For the DistilBART and Flan-T5-Large models, we take $M = 2$ and $K = 2$. For Flan-UL2, Llama-3, DeepSeek-V3, and Granite-3.3, we again use only \texttt{LOO} at the sentence level and do not use \texttt{L-SHAP}. For phrase-level or word-level \texttt{L-SHAP}, we set $M = 1$ and $K = 2$.

\subsection{Baseline attribution methods}
\label{sec:appendix_params:baselines}

\paragraph{\texttt{PartitionSHAP}}
We set the number of model queries (parameter \texttt{max\_evals}) to be approximately equal to the number used by our \texttt{L-SHAP} and \texttt{C-LIME} algorithms. More specifically, since obtaining mixed-level attributions with \texttt{MExGen} requires first performing sentence-level attribution, we add the numbers of queries used during sentence-level and mixed-level attribution. We then take the larger of these two sums for \texttt{L-SHAP} and \texttt{C-LIME} as the number of queries allowed for PartitionSHAP. All other parameters were kept at their default values in the SHAP library.\footnote{\url{https://github.com/shap/shap}, made available under the MIT license}

\paragraph{\texttt{CaptumLIME}} The number of model queries was also set as described above for \texttt{PartitionSHAP}. As discussed in Section~\ref{sec:eval:setup}, the units for attribution are the same as used by \texttt{MExGen} \texttt{C-LIME}. All other parameters were kept at their default values in the Captum library.\footnote{\url{https://github.com/pytorch/captum}, distributed under the BSD 3-Clause license}

\begin{figure*}[th]
    \centering
    \subfloat[\tiny{(\texttt{C-LIME}, DistilBart, XSUM)}]{
        \includegraphics[width=0.24\textwidth]{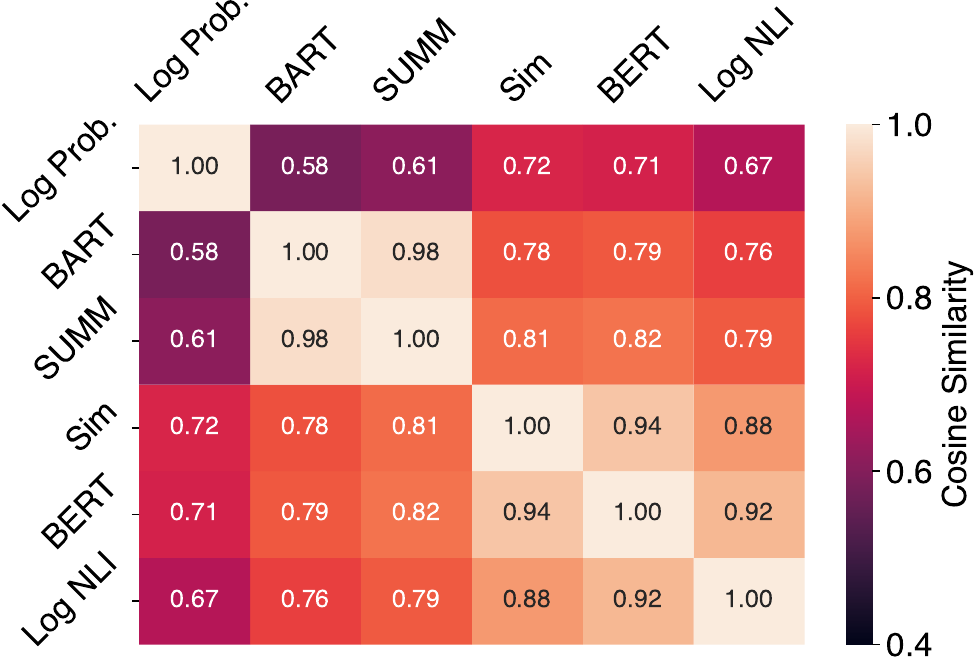}
    }
    \subfloat[\tiny{(\texttt{L-SHAP}, DistilBart, XSUM)}]{
        \includegraphics[width=0.24\textwidth]{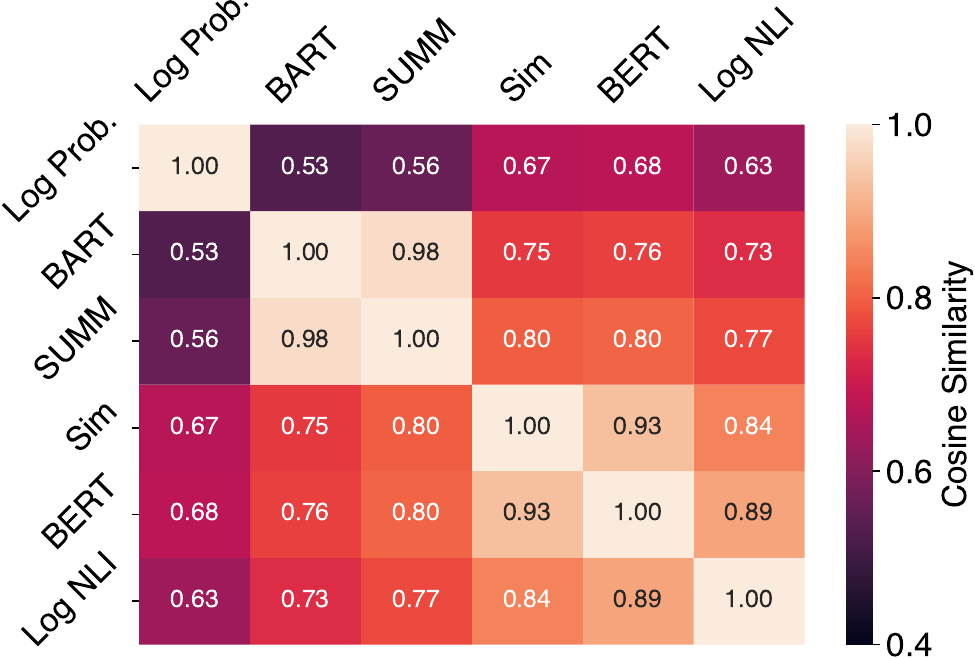}
    }
    \subfloat[\tiny{(\texttt{C-LIME}, DistilBart, XSUM)}]{
        \includegraphics[width=0.24\textwidth]{HeatMaps/LIME_XSUM_spearman.pdf}
    }
    \subfloat[\tiny{(\texttt{L-SHAP}, DistilBart, XSUM)}]{
        \includegraphics[width=0.24\textwidth]{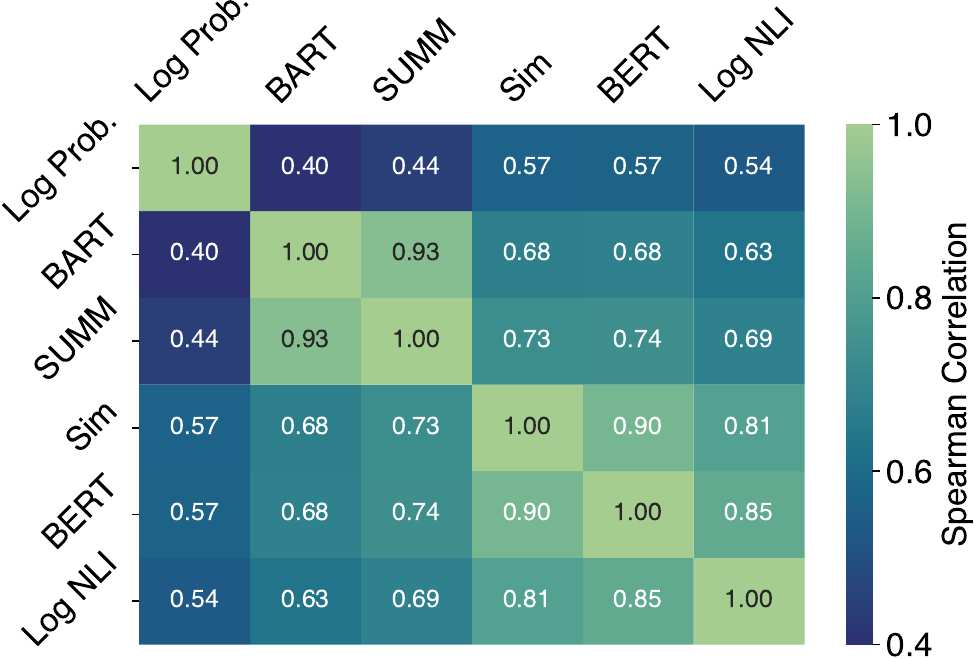}
    }
    \vfill
    \subfloat[\tiny{(\texttt{C-LIME}, Flan-T5, SQUaD)}]{
        \includegraphics[width=0.24\textwidth]{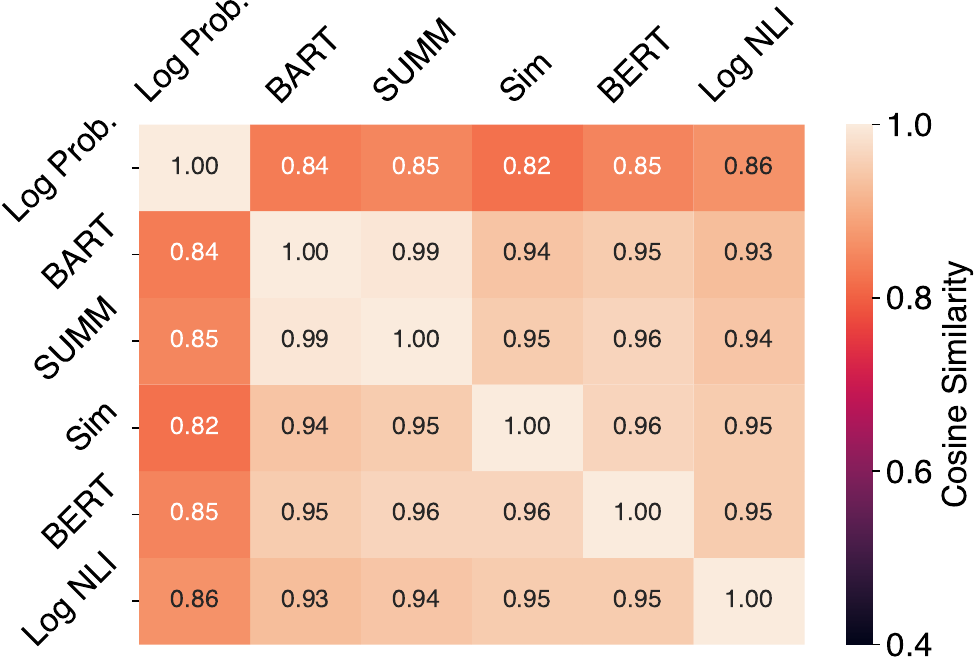}
    }
    \subfloat[\tiny{(\texttt{L-SHAP}, Flan-T5, SQUaD)}]{
        \includegraphics[width=0.24\textwidth]{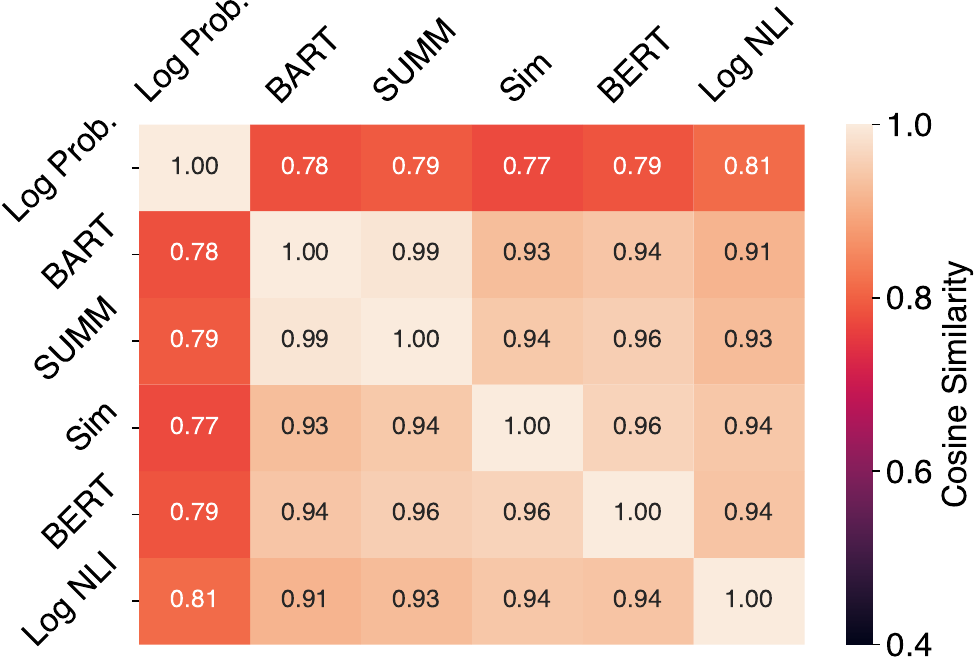}
    }
    \subfloat[\tiny{(\texttt{C-LIME}, Flan-T5, SQUaD)}]{
        \includegraphics[width=0.24\textwidth]{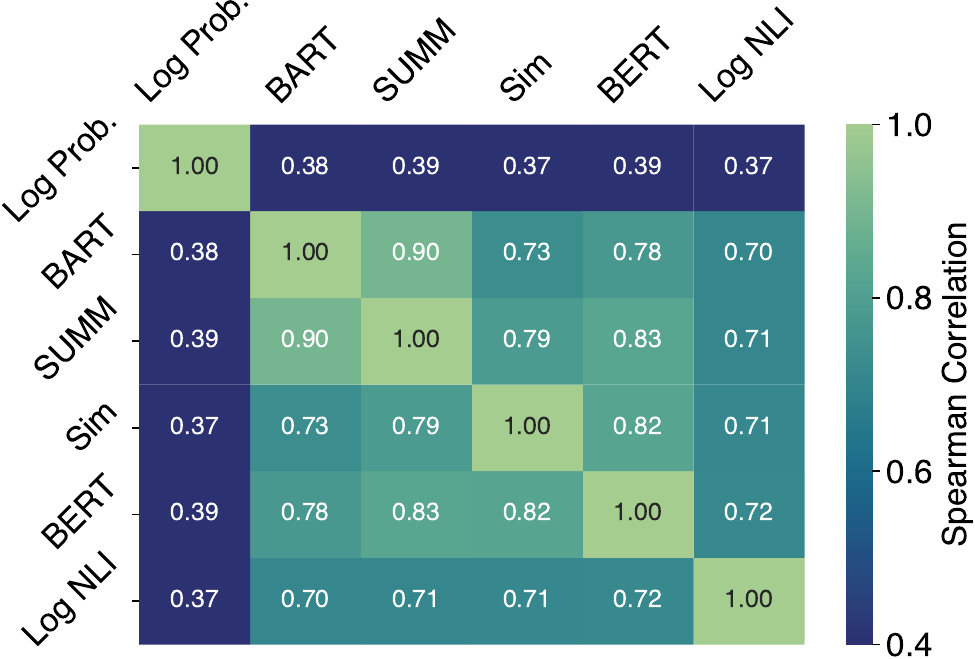}
    }
    \subfloat[\tiny{(\texttt{L-SHAP}, Flan-T5, SQUaD)}]{
        \includegraphics[width=0.24\textwidth]{HeatMaps/SHAP_SQUAD_spearman.pdf}
    }
    \caption{Spearman's rank correlation and cosine similarity for different explanation methods in varying datasets.}
    \label{fig:cosine_similarity_main}
\end{figure*}

\paragraph{LLM Self-Explanation}
Figure~\ref{fig:self_explain_prompt} shows the prompt used to produce LLM self-explanations for the experiment in Section~\ref{sec:eval:self-explain}, where the task is summarization. We first repeat the LLM's summary back to it. We then segment the input article into the same units \texttt{units[0]}, \texttt{units[1]}, ..., (sentence-level or mixed-level) used by \texttt{MExGen}. This controls for input segmentation as done in the comparison with \texttt{CaptumLIME}. To make the ranking task easier, we followed LongBench-Cite \citep{zhang2024longcite} in prepending a numbered tag (<u0>, <u1>, ...) to each unit. The LLM is asked to list only the tags corresponding to units, in order of decreasing importance for generating the summary. The number \texttt{top\_k} of units to be ranked is set to $30\%$ of the total number of units. Since we evaluate perturbation curves up to $20\%$ of the total tokens, the $30\%$ setting provides some margin while not requiring the LLM to rank all units.

The comma-separated output from the LLM is parsed into a list of unit indices. Elements that do not yield an integer or yield an out-of-range integer are dropped (this happened rarely however).

\subsection{Perturbation curves}
\label{sec:appendix_params:perturbation}
\texttt{MExGen} can attribute to mixed units of different lengths in terms of the number of tokens, and these units also differ from those produced by $\texttt{P-SHAP}$. To account for these differences in computing perturbation curves, we consider both the attribution score and the number of tokens for each unit. We rank units in decreasing order of the attribution score divided by the number of tokens (since we plot perturbation curves as functions of tokens perturbed, this ratio can be seen as the slope in the score-tokens plane). Thus, a shorter unit is ranked higher than a longer unit if both have the same attribution score. We then perturb (more precisely remove) the top $k$ units, compute the output score given by the evaluation scalarizer, and increase $k$ until at least $20$\% of the tokens have been removed. 

To average perturbation curves over examples, which have different numbers of tokens, we divide the numbers of tokens perturbed by the total number of tokens in each example to obtain percentages of tokens. We then linearly interpolate onto a common grid of percentages before averaging and computing standard errors.

\subsection{Computing environment} 
Experiments were run on a computing cluster providing nodes with 32 GB of CPU memory, V100 GPUs with 32 GB of GPU memory, and occasionally A100 GPUs with 40 or 80 GB of GPU memory. One CPU and one GPU were used at a time. The total computation time is estimated to be on the order of 1000 hours.

\section{More Automated Evaluation Results}
\label{sec:appendix_automated_evaluation}

\subsection{Scalarizer Evaluation}
\label{sec:appendix_automated_evaluation:scalarizers}

\paragraph{On the Cosine Similarity of Explanations} Figure \ref{fig:cosine_similarity_main} shows the cosine similarity between all the pairs of scalarizers we use.
We show the cosine similarity between the explanations of each method to analyze how aligned the explanations are from different scalarizers.
Each text input $x = x_1, ..., x_d$ receives a multi-level explanation given by $\xi_{S}(x) = (\xi_{S}(x)_1, ..., \xi_{S}(x)_d) \in \mathbb{R}^d$ where each $\xi_{S}(x)_i$ represents the contribution of unit $i$ to the model prediction scalarization computed using the scalarizer $S$.
We define the cosine similarity between scalarizers $S$ and $S'$ as the average of the cosine similarities between the explanations for all available input texts, i.e.,
\begin{equation*}
\texttt{CosSim}(S, S') \triangleq \frac{1}{|X|}\sum_{x \in X} \frac{\langle \xi_{S}(x), \xi_{S'}(x)\rangle}{||\xi_{S}(x)||||\xi_{S'}(x)||}.
\end{equation*}

\begin{figure*}[!t]
    \centering
    \subfloat[\textit{Log Prob} as Evaluation]{
        \includegraphics[width=0.3\textwidth]{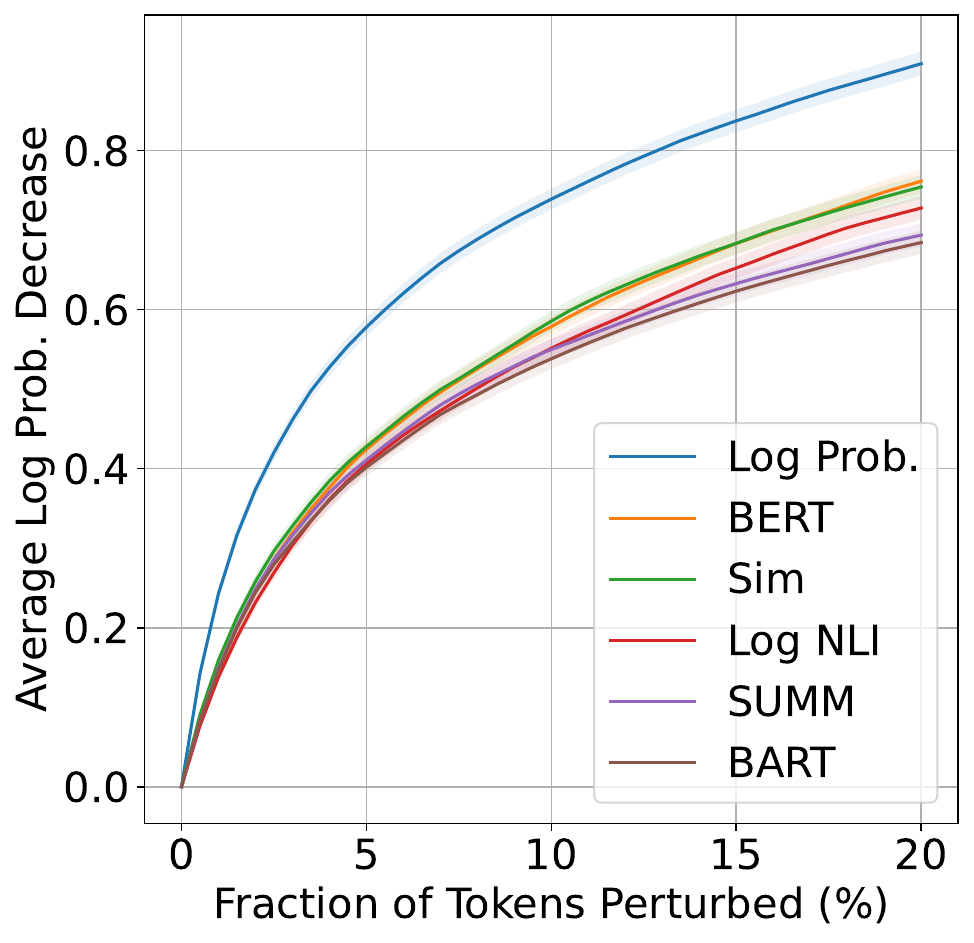}
    }
    \subfloat[\textit{BERT} Score as Evaluation]{
        \includegraphics[width=0.3\textwidth]{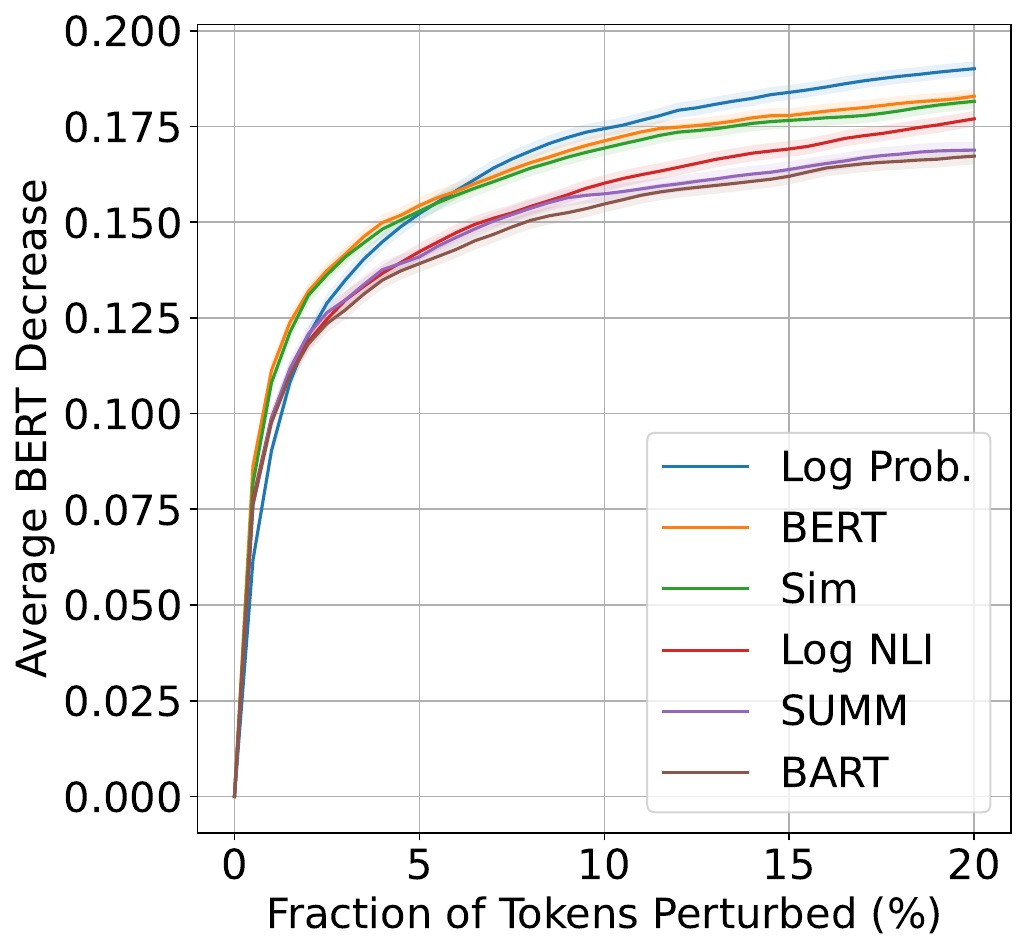}
    }
    \subfloat[\textit{SUMM} Score as Evaluation]{
        \includegraphics[width=0.3\textwidth]{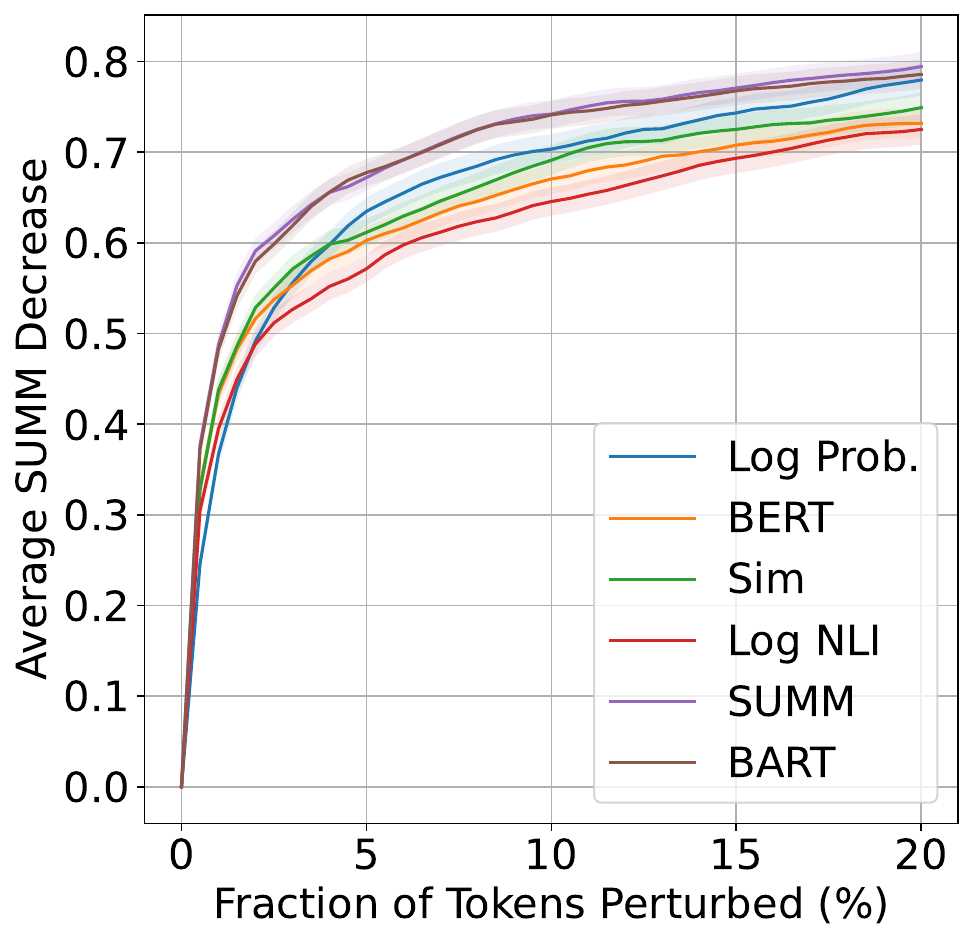}
    }
    \caption{ \label{fig:ScalarCompXSUMDistilbart} 
    Perturbation curves for \texttt{MExGen L-SHAP} with different scalarizers, used to explain the DistilBART model on the XSUM dataset. 
    The curves show the decrease in (a) log probability, (b) BERTScore, and (c) \textit{SUMM} score when removing the most important $p$\% of tokens according to each explanation scalarizer. Shading shows standard error.
    }
\end{figure*}

Figure \ref{fig:cosine_similarity_main} (a) shows the cosine similarities for the scalarizers used by \texttt{MExGen C-Lime} to explain the predictions of the DistilBert model in the XSUM dataset.
Figure \ref{fig:cosine_similarity_main} (b) shows the cosine similarities for the scalarizers used by \texttt{MExGen} \texttt{L-SHAP} to explain the predictions of the Flan-T5-Large model on the SQUaD dataset.

\textit{Similarity Across Scalarizers.} Figure \ref{fig:cosine_similarity_main} indicates that, although some scalarizers lead to similar model explanations, there are occasions where scalarizers are more dissimilar.
Moreover, the similarities between scalarizers not only depend on the scalarizer itself but also on the model and dataset being explained.
For example, Figure \ref{fig:cosine_similarity_main} (b) shows that when using \texttt{MExGen} \texttt{L-SHAP} to provide explanations for the predictions in the XSUM dataset using DistilBrat, the scalarizers \textit{BERT} and \textit{SUMM} are fairly similar ($\texttt{CosSim}(S_{\textit{BERT}}, S_{\textit{SUMM}}) = 0.96$). 
On the other hand, looking at the same pair of scalarizers but for \texttt{MExGen} LIME to provide explanations to the predictions in the SQUaD dataset using Flan-T5-Large, BERT and \textit{SUMM} are more dissimilar ($\texttt{CosSim}(S_{\textit{BERT}}, S_{\textit{SUMM}}) = 0.82$).
This result highlights the necessity of exploring different scalarizers to explain natural language generation, taking into account the task being performed and the main objective of the explanation, i.e., target scalarization.

\textit{Similarity to Logits.} In the SHAP library, \textit{SUMM} is proposed to provide explanations to LLMs that do not provide access to the logits --- hence, the main objective of \textit{SUMM} is to approximate the explanations for the logit when it is not available.
However, Figure \ref{fig:cosine_similarity_main} shows that \textit{SUMM} is not always the best scalarizer for approximating the explanations that would be given if logits were available.
For example, Figure \ref{fig:cosine_similarity_main} (a) shows that the similarity between \textit{SUMM} and logit is near $0.61$.
In contrast, Figure \ref{fig:cosine_similarity_main} (b) shows that the similarity between the explanations generated using \textit{SUMM} as scalarizer has a similarity of $0.78$ with the onex generated using logits.

We are also aware that only comparing the similarities between explanations might not be enough; once, many researchers use the scores to compute the ranking of the importance across all input features (input text units here).
For this reason, next, we compare Spearman's rank correlation to measure the rank stability across different scalarizers.

\textit{Similarity to Log Probability Scalarizer.} In the SHAP library, \textit{SUMM} is proposed to provide explanations when access to logits is not available.
However, Figure \ref{fig:spearman_main_paper} shows that \textit{SUMM} is not always the best scalarizer for approximating the explanations that would be given if logits were available.
For example, Figure \ref{fig:spearman_main_paper} (a) shows the ranking correlation between the scalarizer and the Log Prob scalarizer is higher for \textit{BERT} score.
Figure \ref{fig:spearman_main_paper} (b) shows that the rank generated by both the explanations generated using \textit{SUMM} and \textit{BERT} scalarizer are equally similar to the rank of explanations using the Log Prob scalarizer.

\begin{figure*}[!htb]
    \centering
    \subfloat[Log Prob as Evaluation]{
        \includegraphics[width=0.3\textwidth]{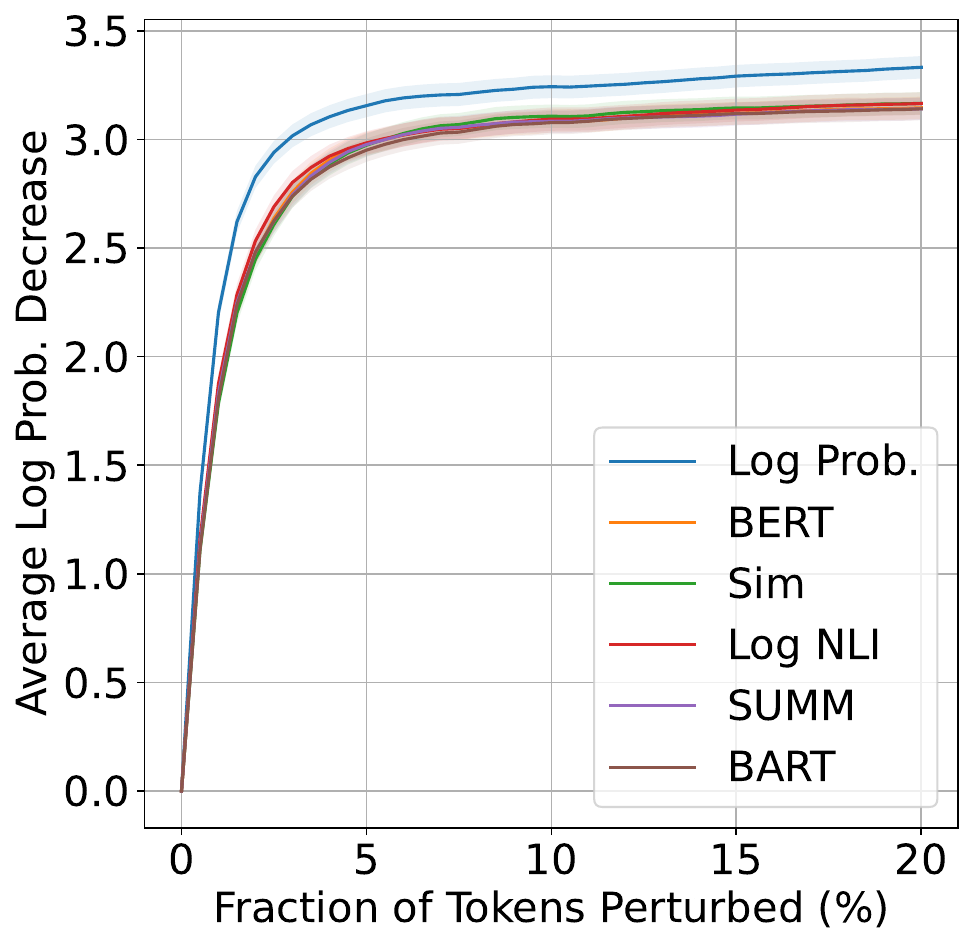}
    }
    \subfloat[\textit{BERT} Score as Evaluation]{
        \includegraphics[width=0.3\textwidth]{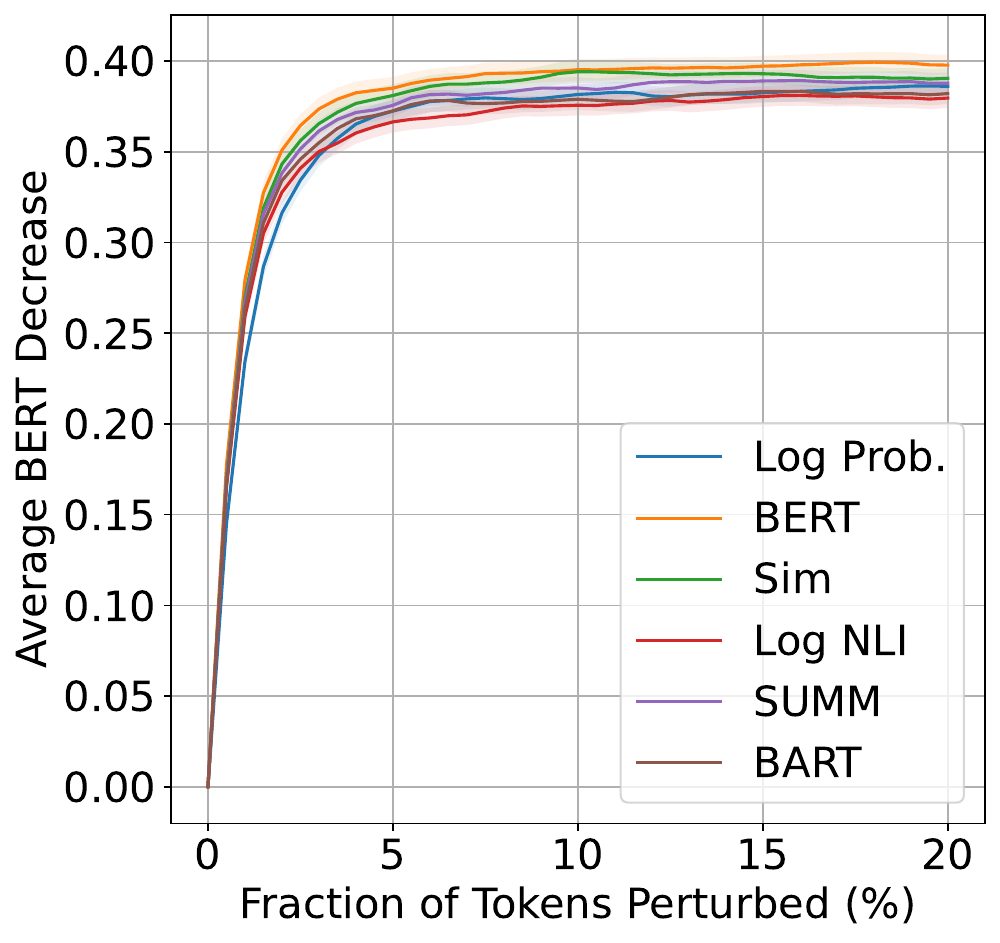}
    }
    \subfloat[\textit{SUMM} Score as Evaluation]{
        \includegraphics[width=0.3\textwidth]{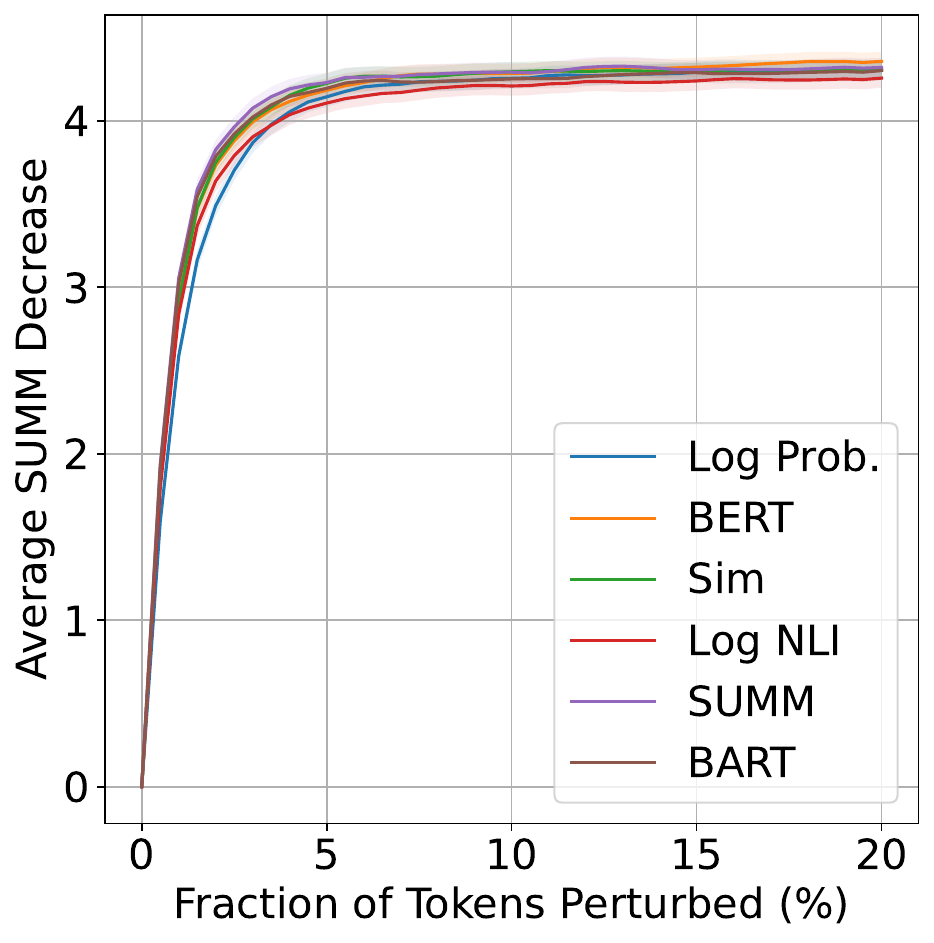}
    }
    \caption{ \label{fig:ScalarCompLimeT5} 
    Perturbation curves for \texttt{MExGen C-LIME} with different scalarizers, used to explain the Flan-T5-Large model on the SQuAD dataset. 
    The curves show the decrease in (a) log probability, (b) BERTScore, and (c) \textit{SUMM} score when removing the most important $p$\% of tokens according to each explanation scalarizer. Shading shows standard error.
    }
\end{figure*}

\begin{figure*}[!htb]
    \centering
    \subfloat[Log Prob as Evaluation]{
        \includegraphics[width=0.3\textwidth]{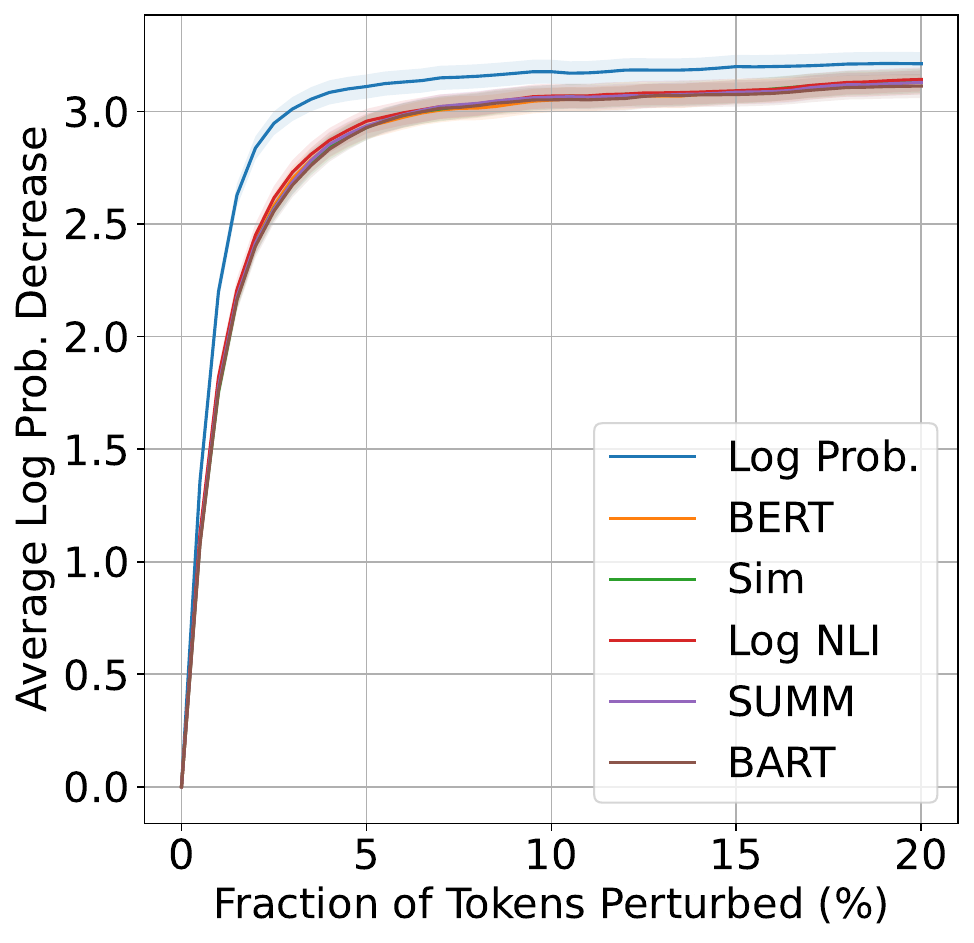}
    }
    \subfloat[\textit{BERT} Score as Evaluation]{
        \includegraphics[width=0.3\textwidth]{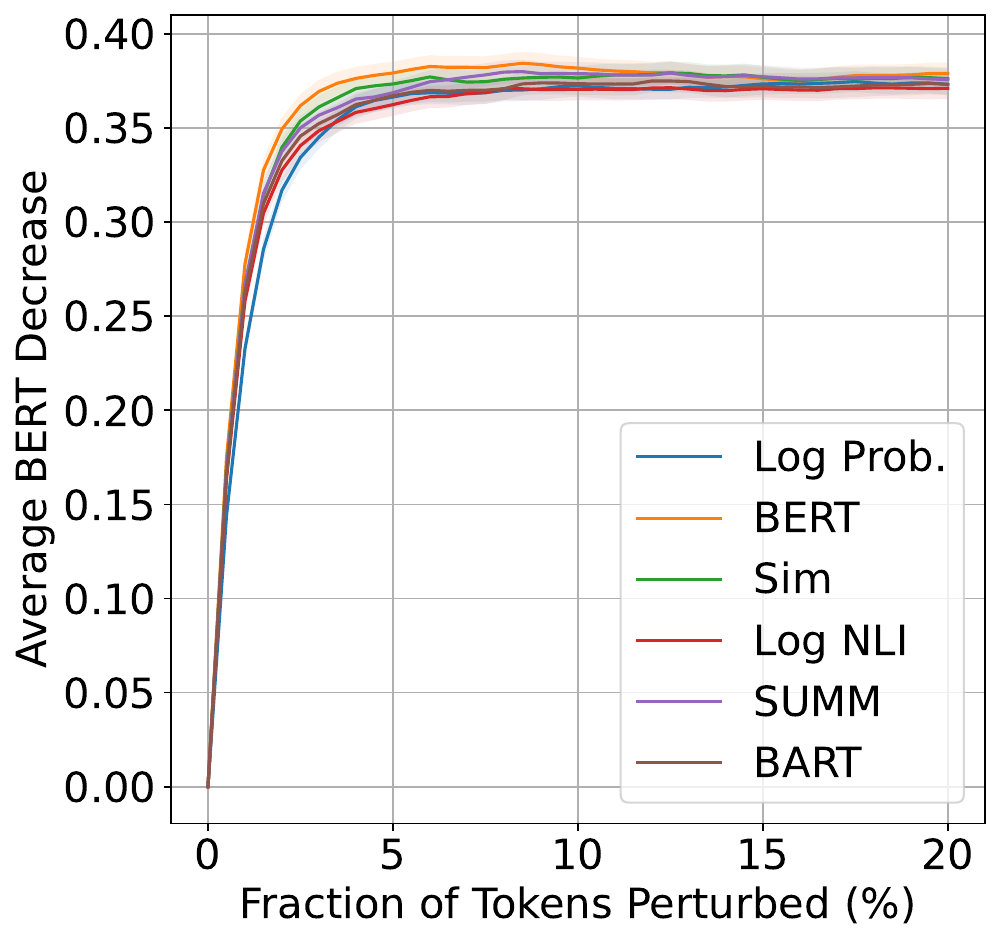}
    }
    \subfloat[\textit{SUMM} Score as Evaluation]{
        \includegraphics[width=0.3\textwidth]{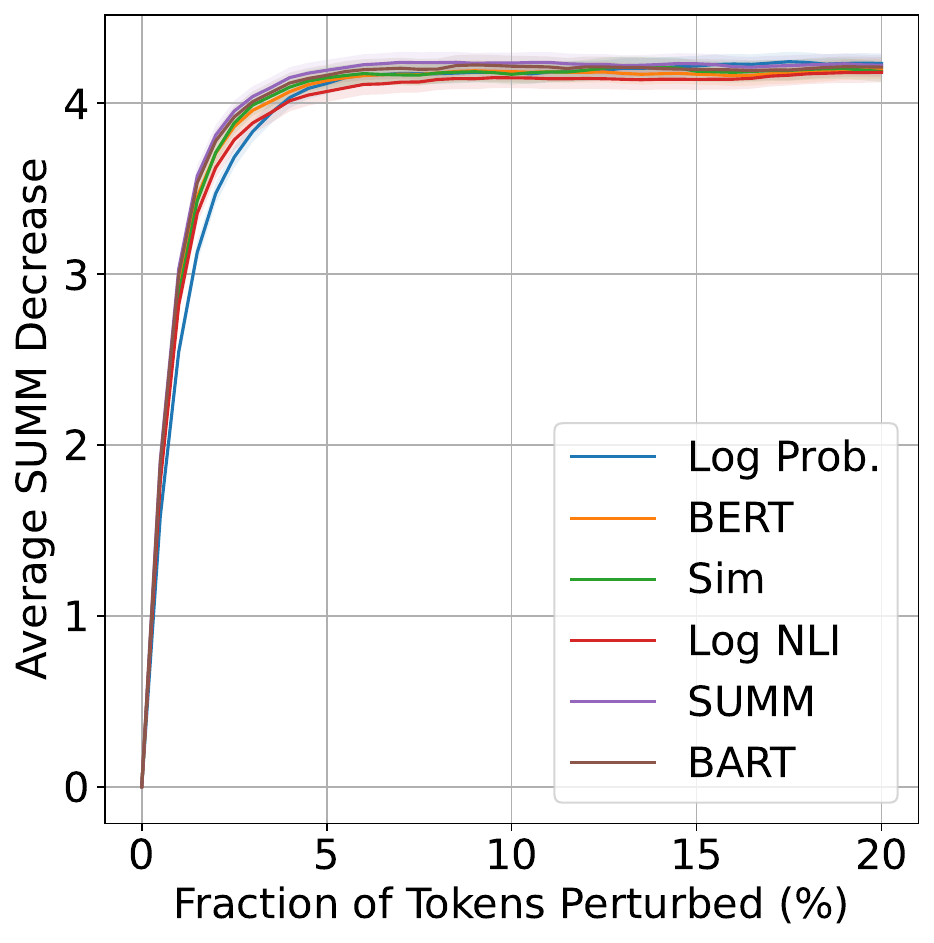}
    }
    \caption{ \label{fig:ScalarCompShapT5} 
    Perturbation curves for \texttt{MExGen L-SHAP} with different scalarizers, used to explain the Flan-T5-Large model on the SQuAD dataset. 
    The curves show the decrease in (a) log probability, (b) BERTScore, and (c) \textit{SUMM} score when removing the most important $p$\% of tokens according to each explanation scalarizer. Shading shows standard error.
    }
\end{figure*}

\paragraph{Perturbation curves for different combinations of scalarizers}
Figures \ref{fig:ScalarCompXSUMDistilbart}, \ref{fig:ScalarCompLimeT5}, \ref{fig:ScalarCompShapT5} show the perturbation curve for different scalarizations.

\newpage

\begin{figure*}[t]
    \centering
    \subfloat[Flan-UL2 on XSUM]{
        \includegraphics[width=0.33\textwidth]{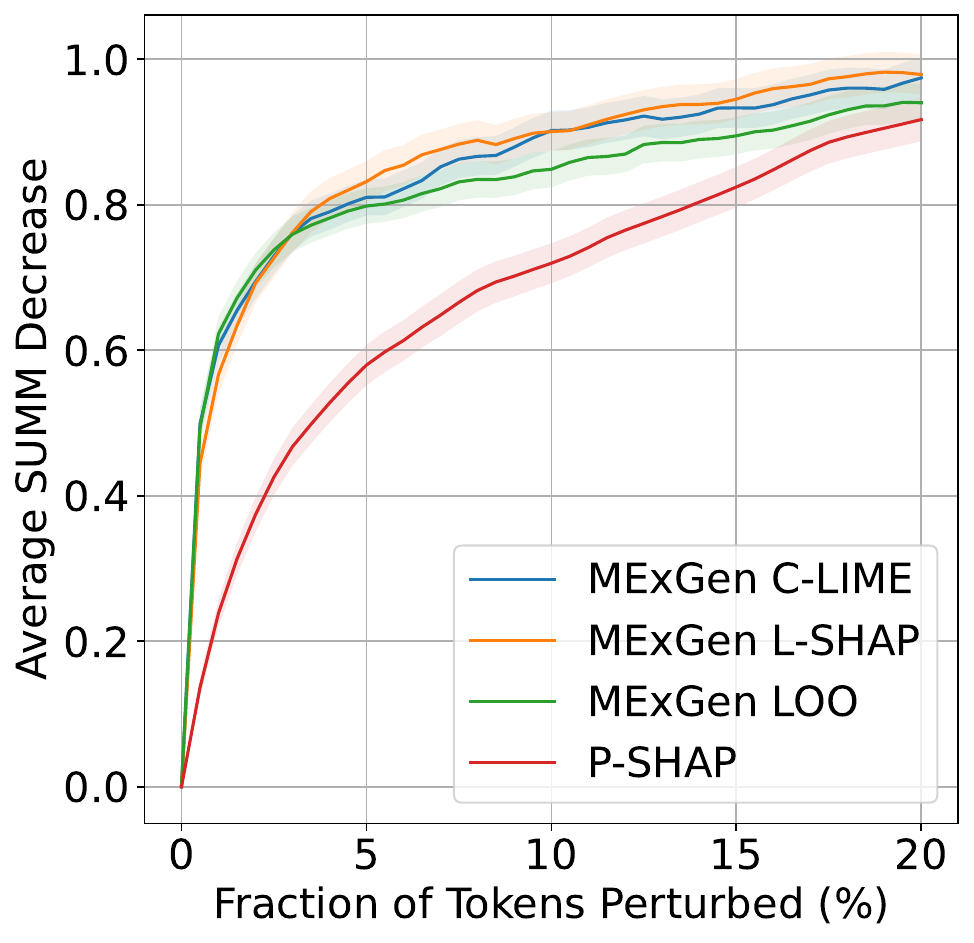}
    \label{fig:Perturbation_flan-ul2_xsum}}
    \subfloat[Flan-UL2 on CNN/DM]{
        \includegraphics[width=0.33\textwidth]{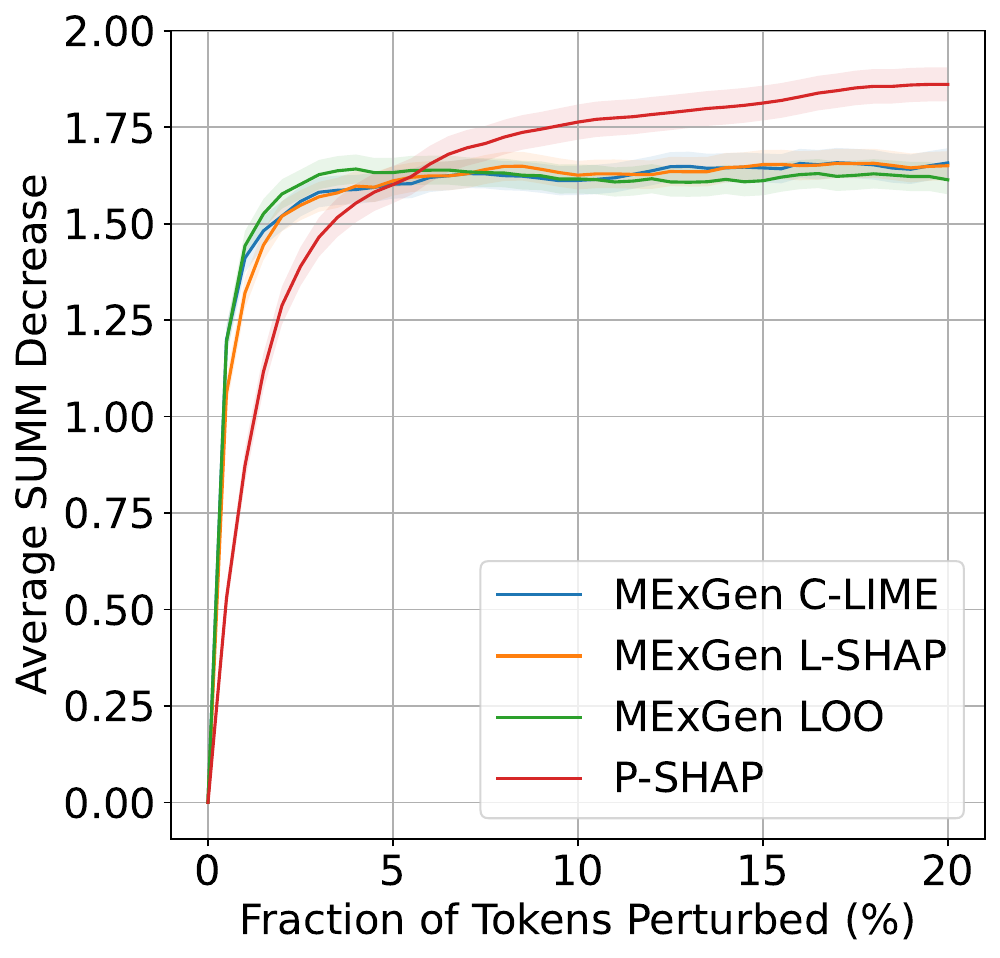}
    \label{fig:Perturbation_flan-ul2_cnndm}}
    \subfloat[Flan-T5-Large on SQuAD]{
        \includegraphics[width=0.33\textwidth]{MethodsComp/Perturbation_flan-t5_squad_1000.pdf}
    \label{fig:Perturbation_flan-t5_squad_2}}\\
    \subfloat[Llama-3 on XSUM]{
        \includegraphics[width=0.33\textwidth]{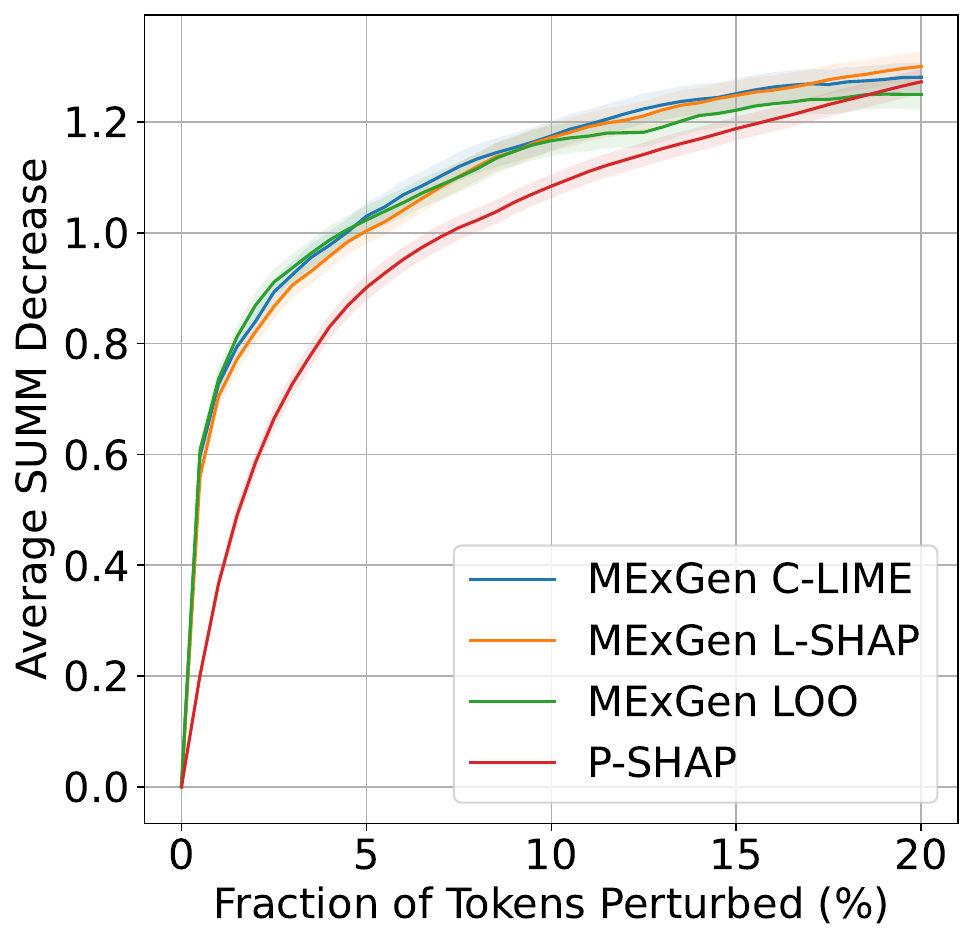}
    \label{fig:Perturbation_llama3-xsum}}
    \subfloat[Llama-3 on CNN/DM]{
        \includegraphics[width=0.33\textwidth]{MethodsComp/Perturbation_llama3-8b_cnndm_500.pdf}
    \label{fig:Perturbation_llama3-cnndm_2}}
    \subfloat[Llama-3 on SQuAD]{
        \includegraphics[width=0.33\textwidth]{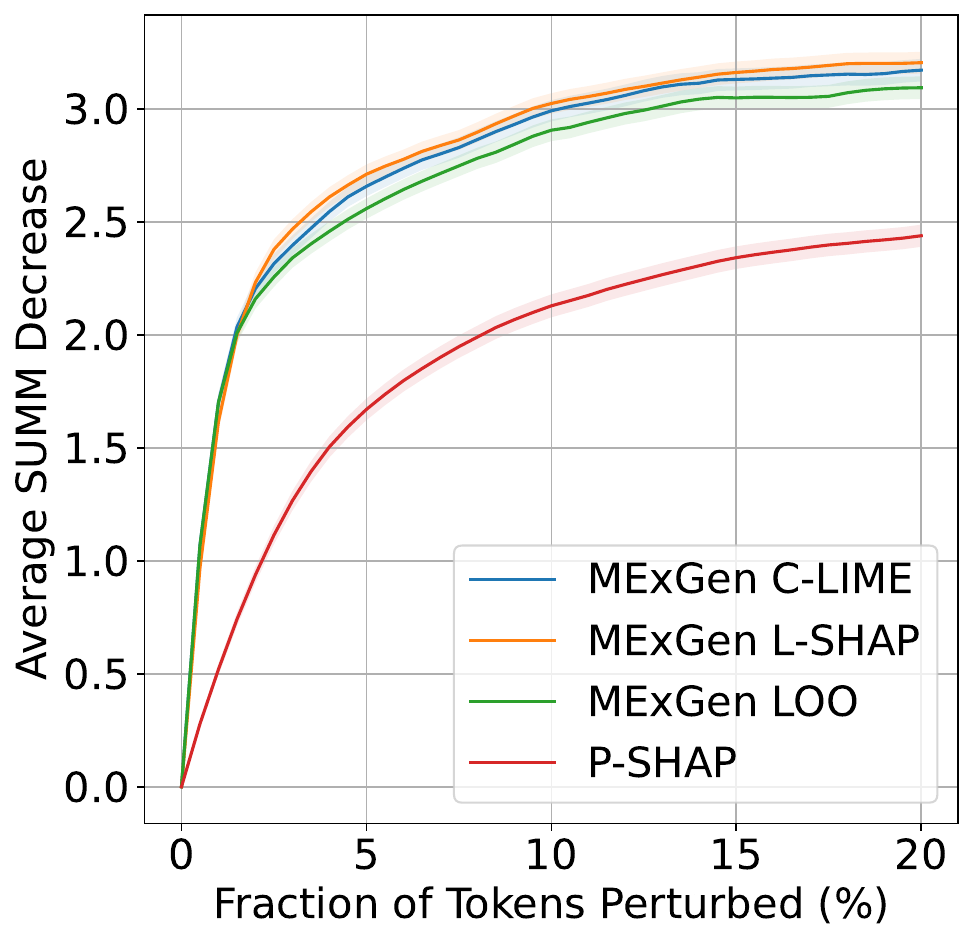}
    \label{fig:Perturbation_llama3_squad}}
    \hfill
    \caption{\label{fig:MethodsComparisonAdd} Perturbation curves (higher is better) from different explanation methods for additional models and datasets: (a) Flan-UL2 on XSUM, (b) Flan-UL2 on CNN/DM, (d) Llama-3-8B-Instruct on XSUM, (f) Llama-3-8B-Instruct on SQuAD, plus (c) Flan-T5-Large on SQuAD and (e) Llama-3-8B-Instruct on CNN/DM repeated from Figure~\ref{fig:MethodsComparison} to facilitate comparison. 
    Shading shows standard error in the mean. 
    }
\end{figure*}

\subsection{Comparison Between Explainers}
\label{sec:appendix_automated_evaluation:explainers}

Figure~\ref{fig:MethodsComparisonAdd} compares the perturbation curves from \texttt{MExGen} instantiations and \texttt{P-SHAP} for additional model-dataset pairs. The patterns are similar to those in Figure~\ref{fig:MethodsComparison}. The one exception is in Figure~\ref{fig:Perturbation_flan-ul2_cnndm} where \texttt{P-SHAP} attains a larger \textit{SUMM} decrease as more tokens are perturbed, but the \texttt{MExGen} curves are still higher for the top 5\% of tokens.

A possible reason for why \texttt{P-SHAP} performs better after a certain fraction of tokens in Figures~\ref{fig:Perturbation_flan-ul2_cnndm}, \ref{fig:Perturbation_flan-t5_squad_2} is as follows: \texttt{P-SHAP} perturbs larger subsets of the input than \texttt{MExGen}, for which we intentionally limit the size of perturbed subsets. These larger subsets may enable \texttt{P-SHAP} to find larger changes in output (higher perturbation curve) at larger fractions of tokens perturbed.

\subsection{Comparison with LLM Self-Explanation}

\begin{figure*}[!htb]
    \centering
    \subfloat[Granite-3.3 on XSUM with \textit{Prob}]{
        \includegraphics[width=0.33\textwidth]{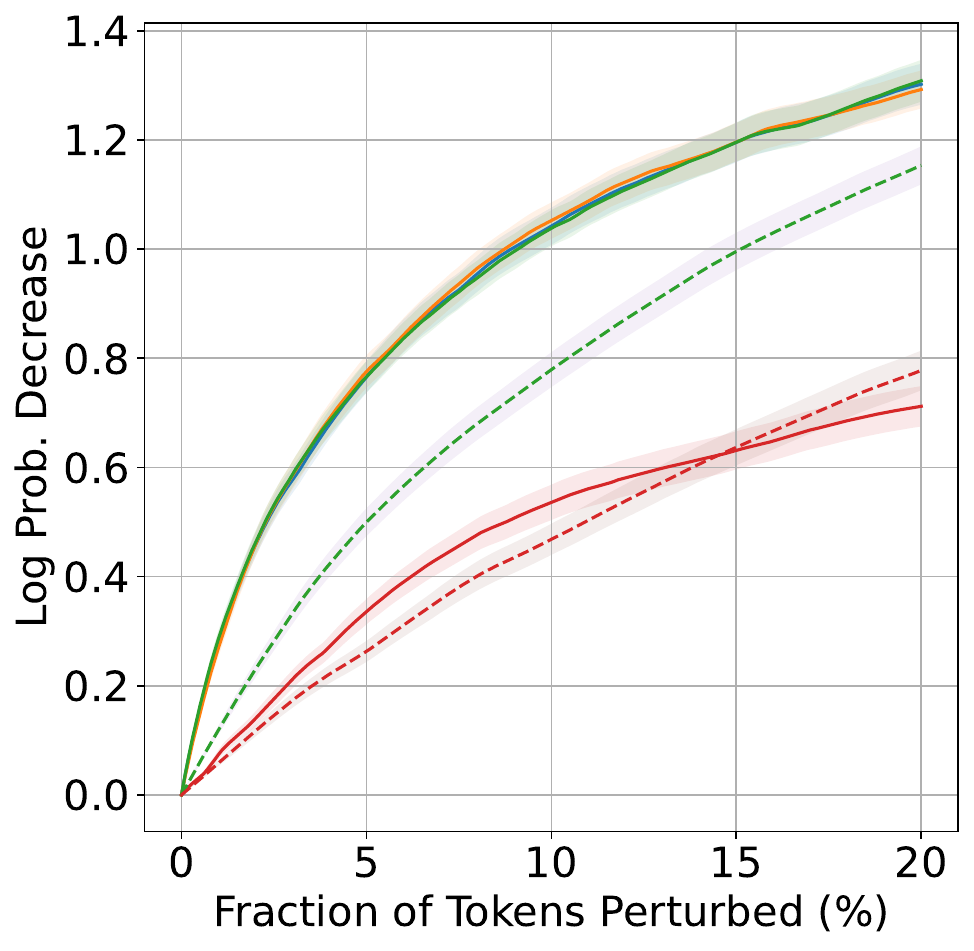}
    \label{fig:self_explain_granite-3-3_prob_xsum}}
    \subfloat[Granite-3.3 on XSUM with \textit{Prob}]{
        \includegraphics[width=0.33\textwidth]{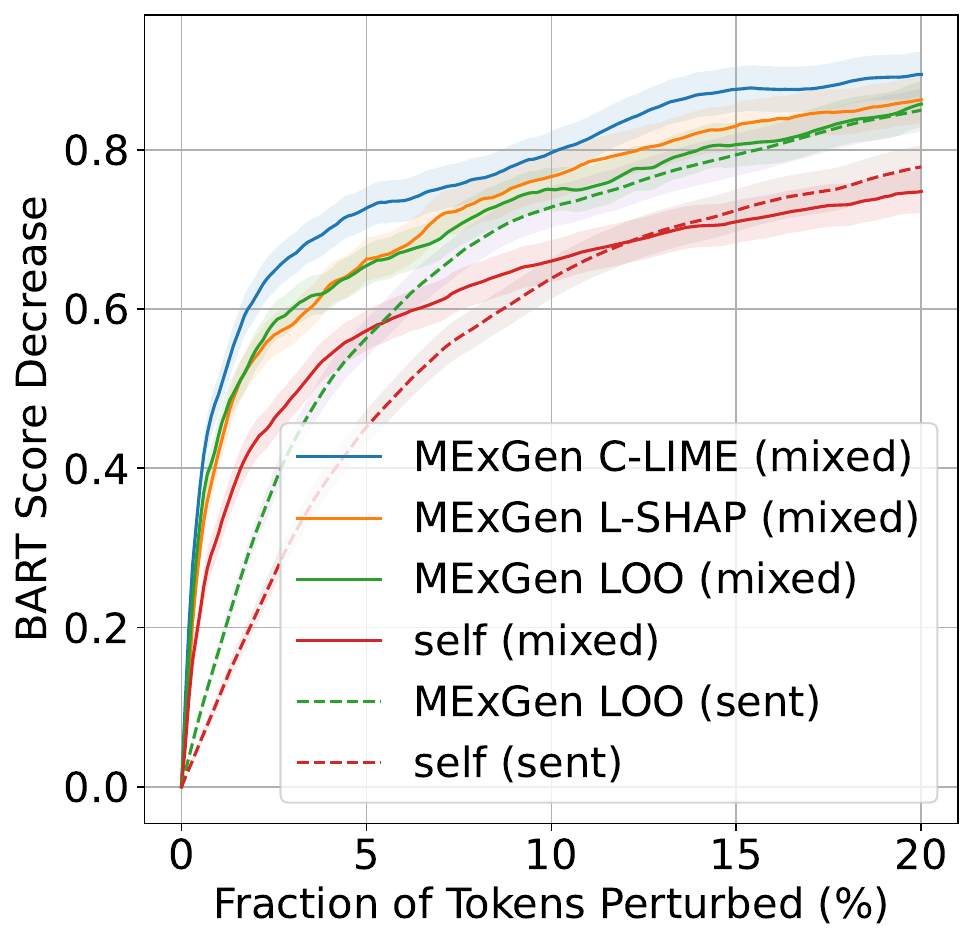}
    \label{fig:self_explain_granite-3-3_text_xsum}}
    \subfloat[DeepSeek-V3 on XSUM with \textit{BART}]{
        \includegraphics[width=0.33\textwidth]{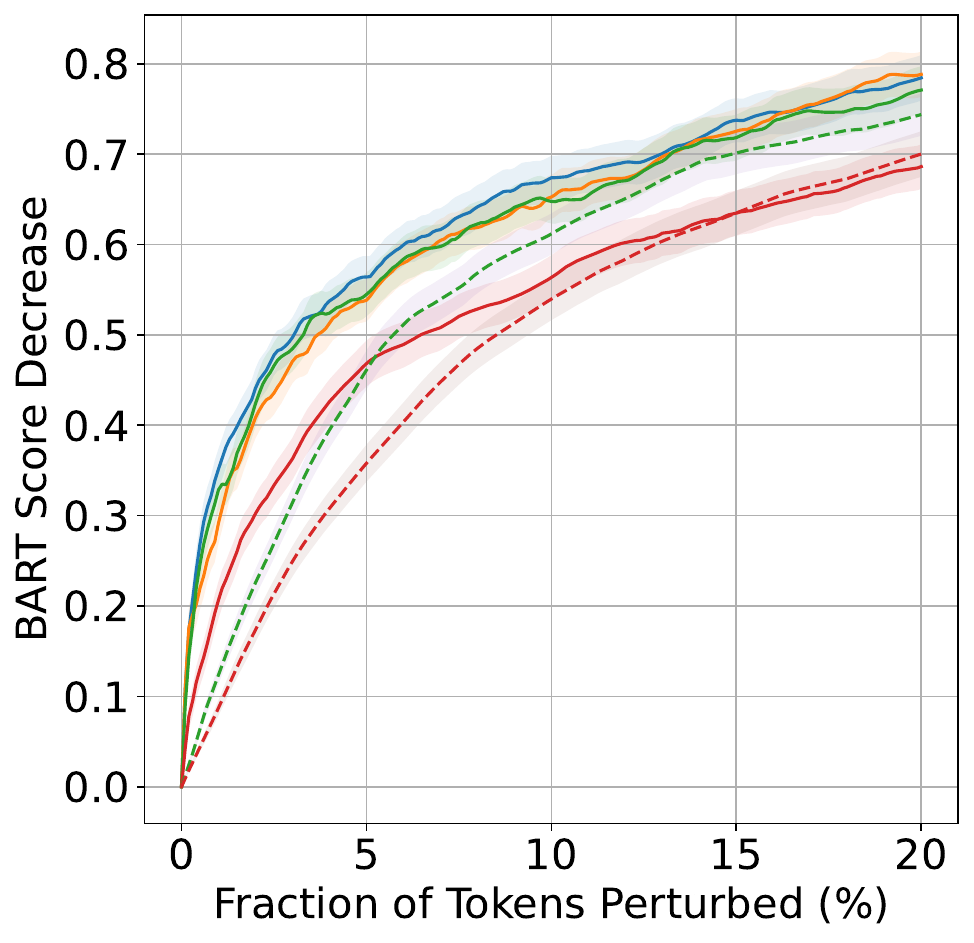}
    \label{fig:self_explain_DeepSeek-V3_text_xsum}}\\
    \subfloat[Granite-3.3 on CNN/DM with \textit{Prob}]{
        \includegraphics[width=0.33\textwidth]{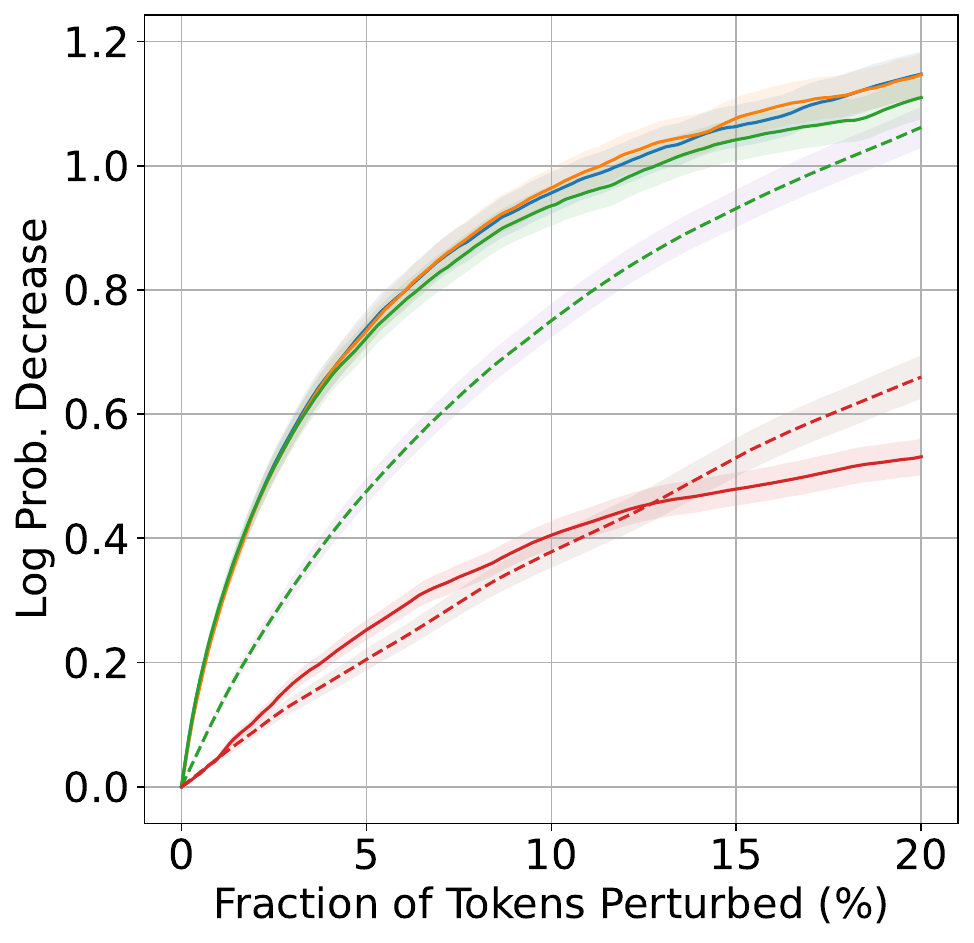}
    \label{fig:self_explain_granite-3-3_prob_cnndm}}
    \subfloat[Granite-3.3 on CNN/DM with \textit{Prob}]{
        \includegraphics[width=0.33\textwidth]{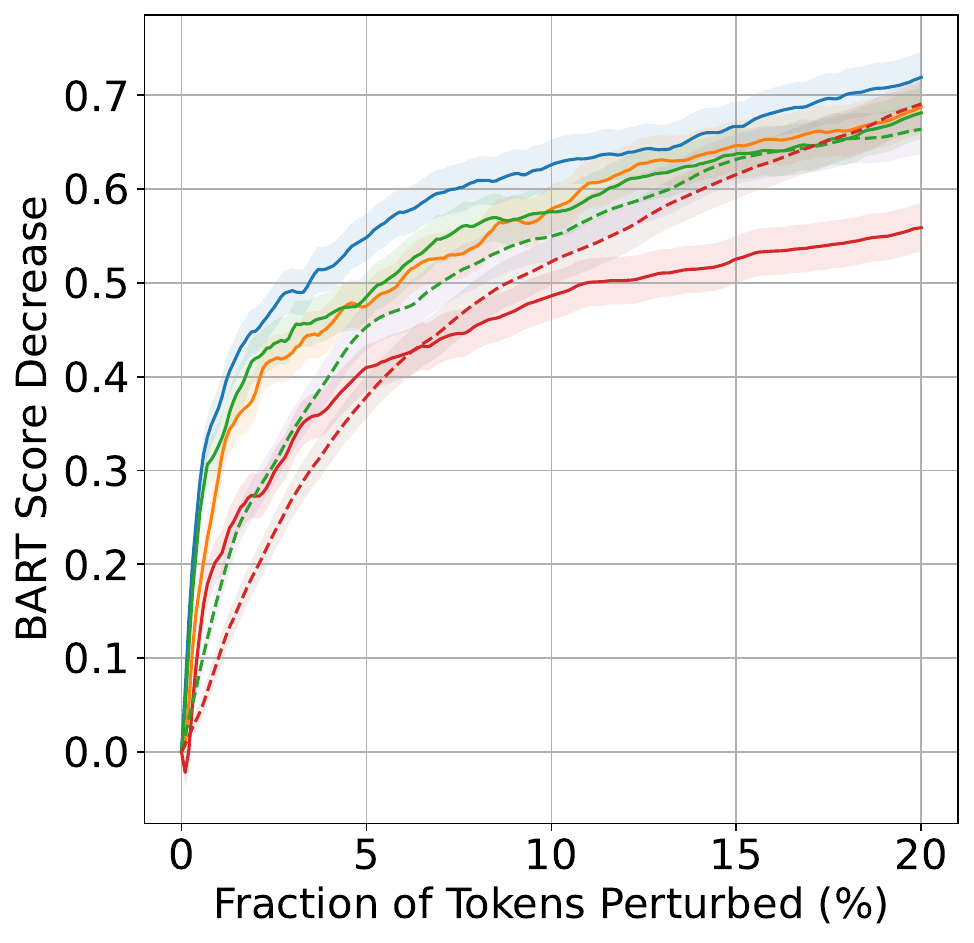}
    \label{fig:self_explain_granite-3-3_text_cnndm}}
    \subfloat[DeepSeek-V3 on CNN/DM with \textit{BART}]{
        \includegraphics[width=0.33\textwidth]{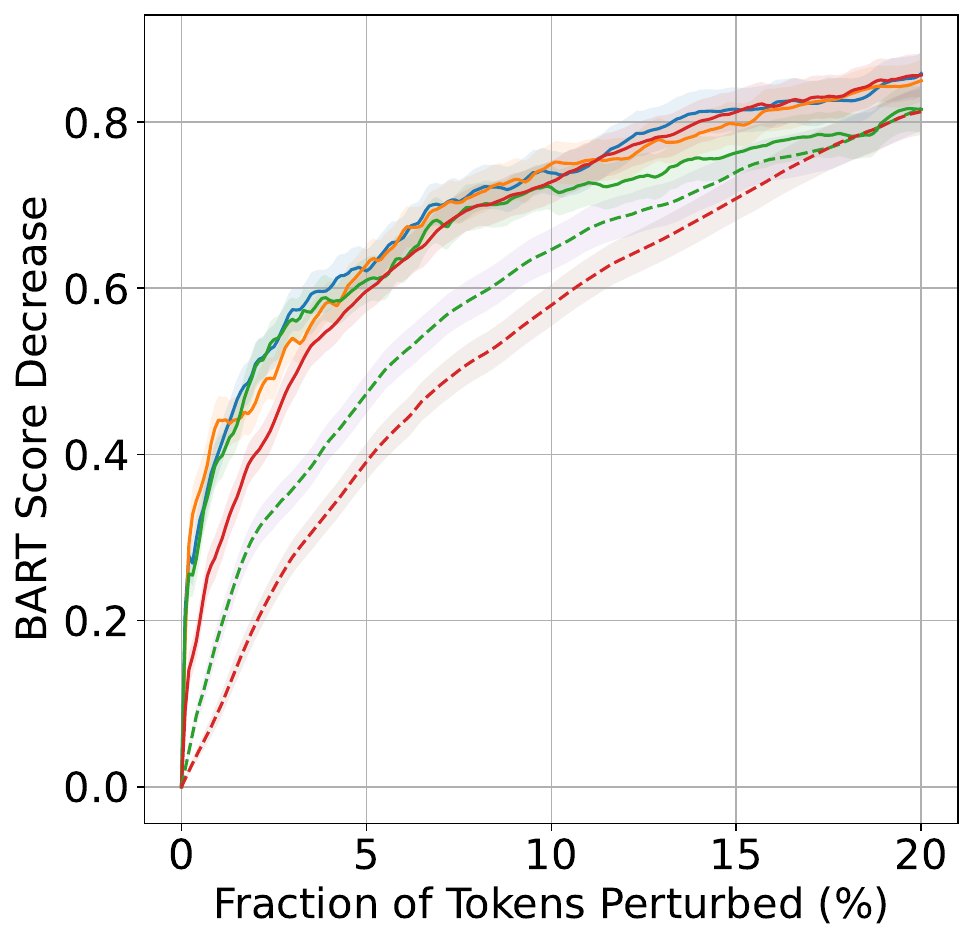}
    \label{fig:self_explain_DeepSeek-V3_text_cnndm}}
    \caption{\label{fig:self_explain} Perturbation curves comparing \texttt{MExGen} variants to LLM self-explanation. LLMs, datasets, and scalarizers (the latter used for both explanation and evaluation): (a) Granite-3.3-8B-Instruct on XSUM with \textit{Prob} scalarizer, (b) Granite-3.3-8B-Instruct on XSUM with \textit{BART} scalarizer, (c) DeepSeek-V3 on XSUM with \textit{BART} scalarizer, (d) Granite-3.3-8B-Instruct on CNN/DM with \textit{Prob} scalarizer, (e) Granite-3.3-8B-Instruct on CNN/DM with \textit{BART} scalarizer, (f) DeepSeek-V3 on CNN/DM with \textit{BART} scalarizer. The legend in (bow) applies to all panels. 
    Shading shows standard error in the mean. 
    }
\end{figure*}

Figure~\ref{fig:self_explain} shows the perturbation curves corresponding to the AUPC values reported in Table~\ref{tab:self-explain}. We also plot the perturbation curve for the intermediate sentence-level attributions (labelled ``\texttt{MExGen LOO} (sent)'') used to obtain the mixed-level \texttt{MExGen} attributions, as well as the curve for the self-explanation using the same sentence-level units (``self (sent)''). The curves closely reflect the AUPC comparison already seen in Table~\ref{tab:self-explain}. For example, \texttt{MExGen} greatly outperforms self-explanation when using the \textit{Log Prob} scalarizer in Figure~\ref{fig:self_explain_granite-3-3_prob_xsum}, \ref{fig:self_explain_granite-3-3_prob_cnndm}. On the other hand, for the text-only \textit{BART} scalarizer and larger DeepSeek-V3 model in Figure~\ref{fig:self_explain_DeepSeek-V3_text_cnndm}, the gap is closed at higher perturbed fractions, but a small gap remains at lower fractions.

\section{User Study}~\label{appendix:user_study}
In this section, we describe our user study. 

\subsection{Participants}~\label{appendix:user_study_participants}
We recruited 96 participants from a large technology company. Those who self-identified as machine learning practitioners using language models were eligible for the study. We filtered out 8 participants who did not pass eligibility checks or did not provide valid responses, resulting in data from 88 participants for our analysis. 

Participation in this study did not involve any significant risks beyond those present in daily life, which we explained in the consent form. 
At the beginning of the study, all participants read about the study procedure, risks, benefits, compensation, and costs, and provided informed consent. They voluntarily participated in the study and were free to withdraw their consent and discontinue participation at any time. Although a formal IRB process does not exist in our institution, we went through an equivalent informal process, including reviewing the study with our peers,
and treated participants in accordance with ethical guidelines for human subjects. 

The study was expected to take about 30 minutes or less. For compensation, each participant received 50 points (a digital currency used within the institution), which was equivalent to \$12.5 USD.

\subsection{Scalarizers}~\label{appendix:userstudy_scalarizer}
Participants perceived \textit{BERT} to be higher in fidelity than \textit{Log Prob.} We ran a binomial test and found that the selection of \textit{BERT} was significantly higher than the random chance ($p<.05$, 95\% CI [.50, .65]). They also preferred \textit{BERT} over \textit{Log Prob} and the choice was statistically significant ($p<.05$, 95\% CI [.56, .71]). The type of attribution methods (e.g., \texttt{C-LIME}, \texttt{L-SHAP}) did not affect the results. 
Participants perceived that the attribution concentration was adequate overall, as the average ratings (\textit{Log Prob}: M=4.32, SD=1.65; \textit{BERT}: M=4.66, SD=1.44) were close to the median of 4 on the 7-point Likert scale. A paired t-test revealed that the difference in the concentration perceptions between scalarizers was not statistically significant.

\begin{table}[h]
\centering
\small
\begin{tabular}{lll}
\hline
Selected Option & \texttt{C-LIME} & \texttt{L-SHAP} \\ \hline
\textit{Log Prob}                & 35.2\%      & 34.1\%        \\ 
\textit{BERT}                & 54.5\%      & 60.2\%        \\ 
Identical                 & 10.2\%      & 5.7\%         \\ \hline
\end{tabular}
\caption{The proportions of participants who selected one of the three options -- \textit{Log Prob}, \textit{BERT}, or `They are identical'. Regardless of attribution methods, significantly more participants chose \textit{BERT} over \textit{Log Prob} when asked to select the one with higher perceived fidelity.}
\end{table}

\begin{table}[h]
\centering
\small
\begin{tabular}{lll}
\hline
Preferred Option & \texttt{C-LIME} & \texttt{L-SHAP} \\ \hline
\textit{Log Prob}                & 29.5\%      & 31.8\%        \\ 
\textit{BERT}                & 62.5\%      & 64.8\%        \\ 
Identical                 & 8\%      & 3.4\%         \\ \hline
\end{tabular}
\caption{The proportions of participants who selected one of the three options -- \textit{Log Prob}, \textit{BERT}, or `They are identical' based on their preference. Regardless of attribution methods,  significantly more participants preferred \textit{BERT} over \textit{Log Prob.}}
\end{table}

\subsection{Attribution Methods}~\label{appendix:userstudy_attribution}
We fitted a Bradley-Terry model~\cite{firth2005bradley} for the outcome of pairwise comparisons between attribution methods. The model computes an `ability' estimate of each method, yielding a complete ranking of methods. Regarding the perceived fidelity, we found that there is a significant difference between \texttt{C-LIME} and \texttt{L-SHAP} ($p<.05/3$ with Bonferroni adjustment), with \texttt{C-LIME} having the highest ability and \texttt{L-SHAP} having the lowest ability. 
The preference data showed the same pattern in which there is a significant difference between \texttt{C-LIME} and \texttt{L-SHAP} ($p<.05/3$ with Bonferroni adjustment), with \texttt{C-LIME} having the highest ability and \texttt{L-SHAP} having the lowest ability. Other pairs of methods were not significantly different. 
Participants perceived that the attribution concentration was adequate overall, as the average rating was close to the median on the 7-point Likert scale (M=4.39, SD=1.47). A repeated ANOVA showed that the differences in perceived concentration among the attribution methods were not significant. 
\begin{table}[h]
\centering
\small
\begin{tabular}{cl}
\hline
Selected vs. Rejected Options  & $p$-value \\ \hline
\texttt{C-LIME}                vs. \texttt{L-SHAP}      & 0.0107 **     \\ 
PartitionSHAP                vs. \texttt{L-SHAP}    & 0.0455 \\ 
\texttt{C-LIME}                 vs. PartitionSHAP      & 0.5691        \\ \hline
\end{tabular}
\caption{There is a significant difference in perceived fidelity between \texttt{C-LIME} and \texttt{L-SHAP}. Significant p-values after Bonferroni adjustment are noted with ** (p<0.05/3).}
\end{table}

\begin{table}[h]
\centering
\small
\begin{tabular}{cl}
\hline
Selected vs. Rejected Options & $p$-value \\ \hline
\texttt{C-LIME}                vs. \texttt{L-SHAP}      & 0.0074 **     \\ 
PartitionSHAP                vs. \texttt{L-SHAP}    & 0.1758 \\ 
\texttt{C-LIME}                 vs. PartitionSHAP      & 0.1758        \\ \hline
\end{tabular}
\caption{There is a significant difference in preference between \texttt{C-LIME} and \texttt{L-SHAP}. Significant p-values after Bonferroni adjustment are noted with ** (p<0.05/3).}
\end{table}

\begin{figure*}[ht]
    \centering
    \includegraphics[width=0.8\textwidth]{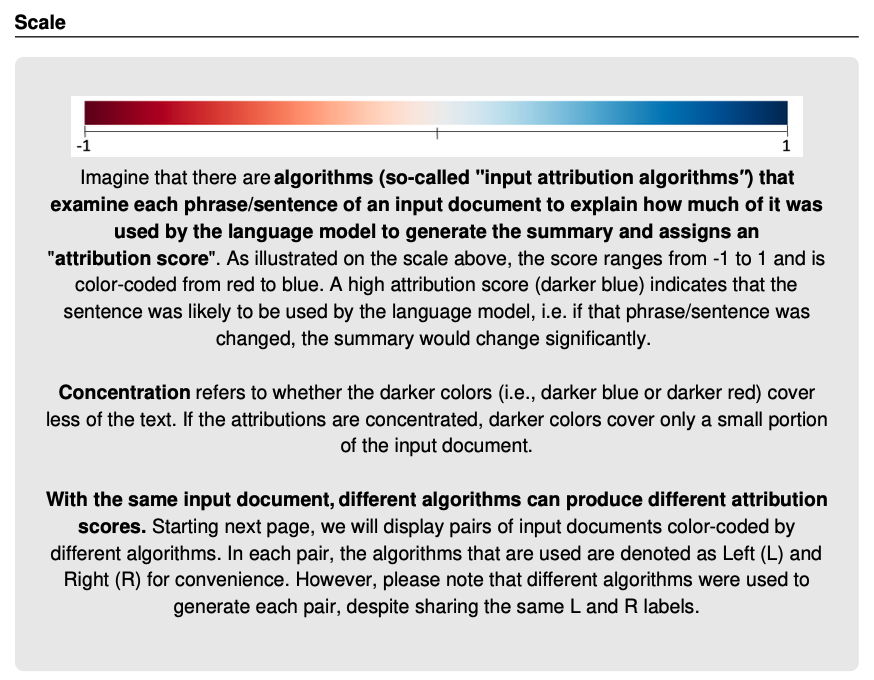}
    \caption{A scale used to color-code the attribution scores.}~\label{fig:scale}
\end{figure*}

\subsection{Granularity Preference}~\label{appendix:userstudy_granularity}
Participants were asked to select their preferred granularity of attributions (sentence-level vs. multi-level). While the number of participants who preferred multi-level granularity (56.2\%) was slightly higher than those who preferred sentence-level granularity (43.8\%), binomial tests indicated that their granularity choice was not statistically significant. The preference for granularity did not significantly vary across attribution algorithms (\texttt{C-LIME}, \texttt{L-SHAP}). 

\begin{figure*}[t]
    \centering
    \includegraphics[width=\textwidth]{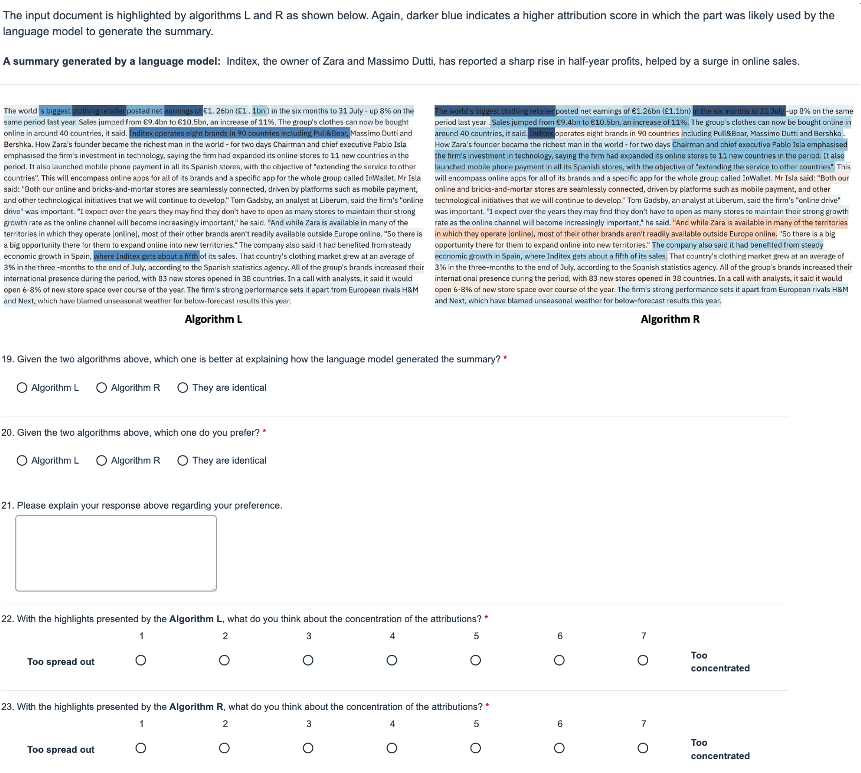}
    \caption{Survey questions. Participants answered a series of questions related to perceived fidelity and general preference comparing a pair of attribution methods, followed by questions related to perceived concentrations of each method.}~\label{fig:interface1}
\end{figure*}

\begin{figure*}[htb]
    \centering
    \includegraphics[width=\textwidth]{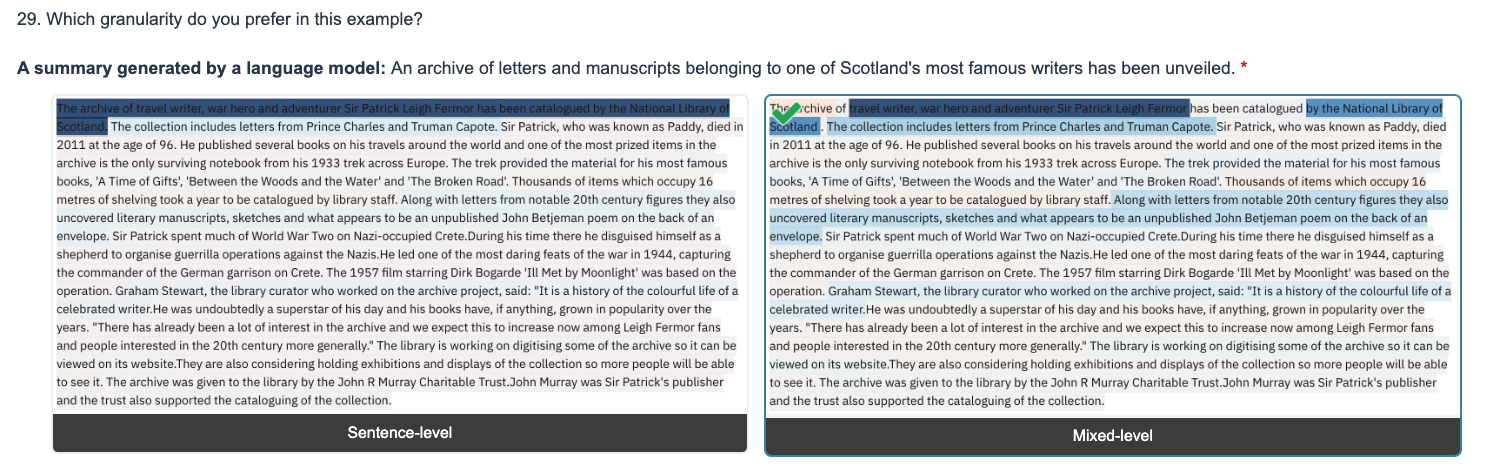}
    \caption{Granularity question. Participants were asked to select their preferred granularity of attributions (sentence-level vs. multi-level).}~\label{fig:interface2}
\end{figure*}

\subsection{Survey}~\label{appendix:UI}
The survey was structured as below with the following instructions and questions:
\begin{enumerate}
    \item \textbf{Consent}
    \begin{itemize}
        \item Select 'I agree' if you are eligible and agree to the terms above. By selecting 'I agree', you give consent to [\textit{Institution Name}] to use your anonymized responses for research and development purposes. You also agree not to provide any information that is confidential or proprietary.
    \end{itemize}
    \item \textbf{Input document (1)}: Suppose you wanted to summarize an input document and used a language model to generate a summary. Please read the text below and answer the following questions. [\textit{A randomly selected input document and a summary were inserted here}]
    \begin{itemize}
        \item Rate the overall quality of the summary. A good summary should be coherent, consistent, fluent, relevant, and accurate. (1: Poor - 7: Excellent)
    \end{itemize}
    \item \textbf{Scale}: Introducing the scale that was used to annotate the attribution scores. See Fig.~\ref{fig:scale}
    \item \textbf{Scalarizer (2 pairwise comparisons)}: Using the selected `input document (1)', same instructions and questions were used as shown in Fig.~\ref{fig:interface1}.
    \item \textbf{Input document (2)}: Suppose you wanted to summarize an input document and used a language model to generate a summary. Please read the text below and answer the following questions. [\textit{Another randomly selected input document and a summary were inserted here}]
    \begin{itemize}
        \item Rate the overall quality of the summary. A good summary should be coherent, consistent, fluent, relevant, and accurate. (1: Poor - 7: Excellent)
    \end{itemize}
    \item \textbf{Attribution algorithms  (3 pairwise comparisons)}: Using the selected `input document (2)', same instructions and questions were used as shown in Fig.~\ref{fig:interface1}.
    \item \textbf{Granularity}: The attribution scores can be presented in two levels of granularity, which are sentence- level and mixed-level granularities. 
    \begin{itemize}
    \item Sentence-level granularity: each sentence in the input document is color coded based on how much of it was used by the language model. 
    \item Mixed-level granularity: each phrase within a few high-scoring sentences is color coded based on how much of it was used by the language model. Low-scoring sentences are coded in the sentence-level granularity. 
    \end{itemize}
    In the next two pages, you will select your preferred granularity for each of the following examples.    
    \begin{itemize}
        \item Which granularity do you prefer in this example? [\textit{A randomly selected summary and two highlighted input documents}] See Figure~\ref{fig:interface2}.
        \item Which granularity do you prefer in this example? [\textit{Another randomly selected summary and two highlighted input documents}]. See Figure~\ref{fig:interface2}.
    \end{itemize}
     \item \textbf{Background}
        \begin{itemize}
        \item What is your job title?
        \item Where is your work location?
        \item What type of work do you do? (Please check all that apply.)
        \item What is your proficiency level in English?
        \item How often do you use language models either as part of your job or as a hobby?
        \item What kind of tasks do you usually do with language models?
        \item Besides the attribution scores/highlights that are shown in this study, what other information would you like to know to help you understand how a language model generated a summary?
    \end{itemize}
\end{enumerate}
Figure~\ref{fig:interface1} and Figure~\ref{fig:interface2} show primary questions we asked in the survey with screenshots.

\section{Future Directions}
\label{sec:future}
\paragraph{Hierarchical explanations} It could be profitable in future work to incorporate the hierarchical explanations discussed in Appendix~\ref{sec:appendix_relWork} into the multi-level \texttt{MExGen} framework. The method of \citet{chen2020lstree} may be especially relevant since it leverages a constituency parse tree to compute word-level importances, which may be related to our use of dependency parse trees.

\paragraph{Word infilling with BERT} We have explored perturbing words by masking them and then calling a BERT model to fill the masks with different words that fit within the sentence. However, we have thus far not seen a quantifiable benefit to using BERT compared to replacing with a fixed baseline value (such as an empty string). Our experience is in line with the the mixed results reported by \citet{pham-etal-2022-double} on using BERT in this manner.

\paragraph{Phrase segmentation} Segmentation of sentences into phrases could of course be done in ways other than our dependency parsing algorithm, for example using constituency parsing instead. A possible advantage of using dependency parsing is that each phrase can be labeled with the dependency label of its root token and treated differently on this basis.

\end{document}